\documentclass{article}

\usepackage{arxiv}

\usepackage[utf8]{inputenc} 
\usepackage[T1]{fontenc}    
\usepackage{hyperref}       
\usepackage{url}            
\usepackage{fancyhdr}
\usepackage{breakcites}
\usepackage{amsmath, amsthm, amssymb, amsfonts}
\usepackage{graphicx}
\usepackage{float}
\usepackage{tabularx}
\usepackage{booktabs}
\usepackage[export]{adjustbox}
\usepackage{multirow} 
\usepackage{caption}
\usepackage{subcaption}
\usepackage[normalem]{ulem}
\useunder{\uline}{\ul}{}
\usepackage{xcolor}
\usepackage{url}
\usepackage{hyperref}
\usepackage{xcolor}
\usepackage{enumitem}
\colorlet{m_red}{red}
\colorlet{m_blue}{black}

\title{Semi-Supervised Deep Learning for Multiplex Networks}

\author{Anasua Mitra \\
\texttt{anasua.mitra@iitg.ac.in} \\
Indian Institute of Technology Guwahati, India \\
\And
Priyesh Vijayan \\
\texttt{priyesh.vijayan@mail.mcgill.ca} \\
McGill University \& Mila \\
\And
Ranbir Sanasam \\
\texttt{ranbir@iitg.ac.in} \\
Indian Institute of Technology Guwahati, India \\
\And
Diganta Goswami \\
\texttt{dgoswami@iitg.ac.in} \\
Indian Institute of Technology Guwahati, India \\
\And
Srinivasan Parthasarathy \\
\texttt{srini@cse.ohio-state.edu} \\
Ohio State University \\
\And
Balaraman Ravindran \\
\texttt{ravi@cse.iitm.ac.in} \\
Robert Bosch Centre for Data Science and AI, \\Indian Institute of Technology Madras, India
}

\begin{document}
\restylefloat{table}
\fancyhead{}
\maketitle

\begin{abstract}
  Multiplex networks are complex graph structures in which a set of entities are connected to each other via multiple types of relations, each relation representing a distinct layer. Such graphs are used to investigate many complex biological, social, and technological systems.  In this work, we present a novel semi-supervised approach for structure-aware representation learning on multiplex networks.  Our approach relies on maximizing the mutual information between local node-wise patch representations and label correlated structure-aware global graph representations to model the nodes and cluster structures jointly. Specifically, it leverages a novel cluster-aware, node-contextualized global graph summary generation strategy for effective joint-modeling of node and cluster representations across the layers of a multiplex network. Empirically, we demonstrate that the proposed architecture outperforms state-of-the-art methods in a range of tasks: classification, clustering, visualization, and similarity search on seven real-world multiplex networks for various experiment settings.
\end{abstract}


\section{Introduction}

Entities in many real-world problems are related to each other in multiple ways. Such relations are often modeled as graph-structured data where the nodes represent entities, and edges between a pair of nodes represent the interactions between the entities. Learning representations for such networked data to mine, analyze and build predictive models has been gaining a lot of traction recently with the advent of deep learning-based network embedding models \cite{wu2020comprehensive, hamilton2020grl}. 

Increasingly such relations are complex, with multiple relationship types linking entities. Such networked data are often naturally represented as multi-layered graphs \cite{kivela2014multilayer}, where each component layer focuses on a specific relation type and can involve different sets of nodes. In this work, we focus on \emph{Multiplex networks}, a special case of multi-layer networks where the graphs in all the layers share the same set of nodes with distinct relations in different layers. Such multiplex network structures are observed in numerous environments such as bibliographic networks, temporal networks, traffic networks, brain networks, protein-drug-disease interaction, etc. The involvement of the same set of nodes across multiple types of relations, distinctive structures in different layers, and the interplay among various layers of networks --- makes representation learning of multiplex networks a challenging task. 

Existing multiplex Network Representation Learning (NRL) methods learn node embeddings that encode the local relational structure of nodes by using graph convolutions \cite{ghorbani2019mgcn, luo2020deep, park2020unsupervised} or random walks \cite{liu2017principled, zhang2018scalable} within a subgraph centered at the node of interest. Though there are many powerful models to learn local structures, only a few works encode global structures \cite{luo2020deep, park2020unsupervised} even in the case of the more widely researched simple homogeneous graphs. Global structural information is encoded in representation learning models through one of three approaches: (i) clustering constraints \cite{wang2017community, tian2014learning}; (ii) auto-encoding objectives on the adjacency matrix \cite{wang2016structural, wang2017community, tian2014learning} or node embeddings \cite{gao2019graph}; and (iii) Mutual Information Maximization (InfoMax)~\cite{velivckovic2018deep, sun2019infograph} objectives that maximize the Mutual Information (MI) between the representations of local nodes and the global summary of the graph derived from the local contexts of all the nodes \cite{velivckovic2018deep, sun2019infograph, park2020unsupervised, ren2020hdgi}. 

 Clustering constraints are also often realized with auto-encoding objectives, that in general, are challenging to scale \cite{wang2017community, tian2014learning}. In contrast to the first two methods, the InfoMax based approaches use Graph Neural Networks (GNNs) to obtain both local and global context and are potentially more scalable \cite{velivckovic2018deep, park2020unsupervised, ren2020hdgi}.  
However, InfoMax objectives that encode global information assume a shared global graph context for all the nodes despite the fact that, in most cases, every node has a different global structure rooted at each node.
This calls for a different contextualized global graph representation for each node (analogous to the notion of personalization). 
\smallskip
\\\textbf{Contributions.} In this work, we propose the first node-contextualized InfoMax based semi-supervised learning architecture for multiplex networks. The primary contributions of our work are:
\begin{itemize}[leftmargin=*]
    \item Motivated by the need for contextualized global graph representations, we propose a novel {\it joint} node and cluster representation learning model that defines a structure-aware intra-layer graph context for a node. 
    \item In the semi-supervised setting, the cluster constraints are provided by the partial label information and are shared across layers to learn similar clusters across relational layers in terms of label correlations. To facilitate this, we further constrain the nodes connected by cross-edges to have similar embeddings, thereby indirectly influencing the layerwise InfoMax objective to capture global cluster information across multiple layers.
    \item We evaluate the model on seven multiplex networks for node classification, clustering, and similarity search. Our proposed model achieves the best overall performance outperforming state-of-the-art methods like DMGI~\cite{park2020unsupervised}, MGCN~\cite{ghorbani2019mgcn}, HAN~\cite{wang2019heterogeneous}. 
    \item Also, the learned node embeddings lead to well-separated homogeneous clusters in t-SNE visualizations.
\end{itemize}

\section{Background and Key Intuition}
\subsection{Notation and Problem Statement} \label{subsection:formalization}
\textbf{Notation:} Let the multi-layer representation of a multiplex network with vertex set, $\mathcal{V}$ and relation set, $\mathcal{R}$ such that $|\mathcal{R}| > 1$ be defined by an $|\mathcal{R}|$-layered graph, $\mathcal{G} = \{\mathcal{G}_1, \mathcal{G}_2, ..., \mathcal{G}_{|\mathcal{R}|}\}$, where $\mathcal{G}_r$= $(\mathcal{V}, A_r)$ with $A_r$ being the adjacency matrix for the $r$\textsuperscript{th} layer corresponding to the $r$\textsuperscript{th} relation. We generalize the adjacency matrix notation to include both intra-layer and inter-layer edges between nodes in different layers by letting $A_{(r,r)}$ to denote $r$\textsuperscript{th} layer's adjacency matrix corresponding to its intra-layer edges and $A_{(r,s)}$ to denote the inter-layer edges between layer $r$ and $s$ when $r\neq s$. Note that $A_{(r,r)}, A_{(r,s)} \in \mathbb{R_+}^{|\mathcal{V}|\times|\mathcal{V}|}$ as all the layers share the same set of nodes, $\mathcal{V}$. Often, the nodes are associated with a feature set, $\mathcal{F}$ and the node feature matrix is denoted as $X \in \mathbb{R}^{|\mathcal{V}|\times|\mathcal{F}|}$.

\noindent \textbf{Semi-Supervised Learning Task:} Given a multiplex graph, $\mathcal{G}=(\mathcal{V}, A, X)$, a label set, $\mathcal{Q}$ and set of labeled nodes, $\mathcal{L}$ with ground truth label assignment matrix, $Y\in \{0,1\}^{|\mathcal{V}|\times|\mathcal{Q}|}$, the task is to predict labels for all unlabeled nodes, $\mathcal{U} = \mathcal{V} \setminus \mathcal{L}$. For efficient Semi-Supervised Learning (SSL) on multiplex networks, it is essential to learn a low $d$-dimensional $(d \ll |\mathcal{F}|)$ node embedding, $Z \in \mathbb{R}^{|\mathcal{V}|\times d}$ that encodes relevant structural and label correlation information within and across layers, useful for downstream machine learning tasks such as node classification and clustering.

\subsection{Multiplex NRL and InfoMax Objective}
\noindent{\bf Node Representation Learning:}
Multiplex Network Representation Learning methods encode useful information for all the nodes into a low $d$-dimensional node embedding, $Z_r  \in \mathbb{R}^{|\mathcal{V}|\times d}$ for each layer $r$ and then aggregate information across layer by leveraging the cross edges, into a joint embedding, $Z \in \mathbb{R}^{|\mathcal{V}|\times d}$. 

Graph Convolutional Networks (GCNs)~\cite{kipf2016semi, schlichtkrull2018modeling} are widely used node embedding architectures that encode attribute-based local structural information from a node's multi-hop neighborhood. In the context of multiplex networks, GCNs~\cite{zitnik2018modeling, velivckovic2018deep, park2020unsupervised} are used layer-wise to obtain node embeddings based on the intra-layer edges. Then, to get a joint node embedding, embeddings from different node counterparts across layers are aggregated via the cross-edges. 
\vspace{-1em}
\begin{figure*}[ht] \centering
\minipage{0.75\textwidth}
  \includegraphics[width=\linewidth, keepaspectratio]{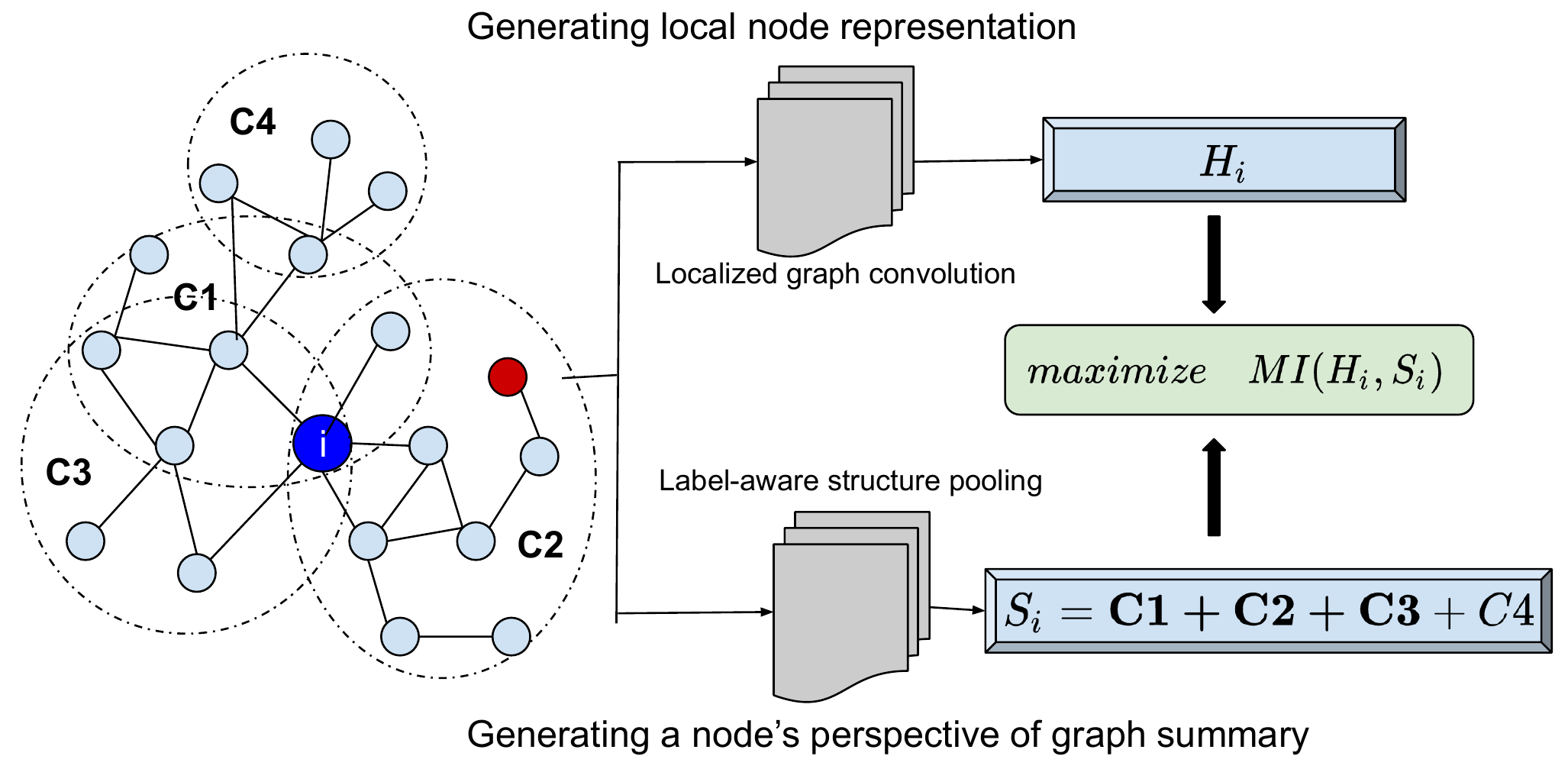}
\endminipage \caption{Label correlated structure-aware InfoMax}
\label{figure:intuition}
\end{figure*}
\vspace{-1em}
\noindent{\bf Encoding Global Information with InfoMax based objective:}
While GCNs are powerful models to encode local structures, they do not encode global contextual information. On that front, recent efforts in the NRL community have adopted the Mutual Information Maximization (InfoMax) objective, initially proposed in image feature extraction pipelines for learning structurally dependent rich local and global representations of images --- to the graph domain to learn rich node \cite{velivckovic2018deep, park2020unsupervised} and graph-level representations \cite{sun2019infograph}. 
In problems where the data is sampled from a set of graphs, each data instance is a graph, and the task is to learn a global graph representation wherein the InfoMax based models learn graph representations by maximizing their mutual information (MI) with the local node representations \cite{sun2019infograph}. In this work, we are interested in the semi-supervised transductive setting. Given a partially labeled graph, we look to learn node representations that allow us to predict labels for the rest of the nodes in the same graph. In this setting, existing Infomax models learn the local node representations by maximizing its MI with a single global graph representation \cite{velivckovic2018deep, park2020unsupervised}. As we argue in the next section, where we describe this work's intuition, this single global representation is often inadequate. 

From the computational aspect, the mutual information between two variables can be maximized by leveraging the KL-divergence between their Joint distribution and the product of marginals. However, since estimating MI in high dimension and continuous data is complex, in practice, scalable Neural MI estimators \cite{belghazi2018mine} that maximize a tractable lower-bound are used. {\it Noise Contrastive Estimation-based (NCE) loss} that discriminates samples from a true joint distribution against a noisy product of marginals (negative samples) are simple yet effective ways to realize this lower bound. It can be viewed from the predictive coding perspective, where given a global-whole representation, the task is to predict a corresponding local-part representation. This forces the discriminator to provide a high score to a related pair of local-global representations compared to unrelated pairs. Henceforth, unless specified otherwise, we adopt minimizing the NCE loss for this purpose. 

\subsection{Key Intuition} \label{sec:intuition}
In a typical InfoMax based NRL setup, the global context is defined by all nodes in the graph. Thus, each node in the graph does not have its own contextual view of the graph. Instead has a shared global context that is the same for all the nodes, {\it even though they may be structurally connected differently within the graph}. For example, from the graph in Fig: 1, the red node and blue node belonging to the same cluster C2 will have different non-local network measures such as betweenness measure, participation coefficient \cite{guimera2005functional}, etc., as the structure of the (sub)graph centered around these nodes are differently connected to the rest of the graph since one is in the center of a cluster, and other is at the end of a whisker.

When a naive global graph summary function such as the average of all node embeddings is used, the global context for all nodes becomes the same as their global context is isomorphic. Naively maximizing the MI of a nodes' local representations with a shared global context might bias the model to encode trivial and noisy information found across all the nodes' local information. Albeit, naively defining a different global context for each node, such as a sub-graph-based approach, will shoot down the original objective of learning useful shared information from across the graph. 

Thus, this calls for a careful design of a \textit{contextualized representation of shared global information} that facilitates encoding relevant non-trivial shared information present across the graph when maximizing the MI with the local node information. In Figure~\ref{figure:intuition}, even though the example graph has many clusters, only $C1, C2, C3$ are relevant to the blue node $i$. Therefore, the global context for node $i$ should be more inclined towards $C1, C2, C3$ instead of a naive summary of all candidate clusters. In light of this simple intuition and motivated by participation scores, we propose a cluster-based InfoMax objective to learn node representations. \textit{The clusters encode shared global graph information, and the node-specific global context is obtained by aggregating information from the clusters with which the node is associated.} In particular, for the semi-supervised classification task, we define label-correlated structure-aware clusters that jointly learn node and cluster representations by optimizing the InfoMax principle. 
\section{Proposed Methodology} \label{section:proposed}
\begin{figure*} [ht] \centering
\minipage{0.88\textwidth}
  \includegraphics[width=\linewidth, keepaspectratio]{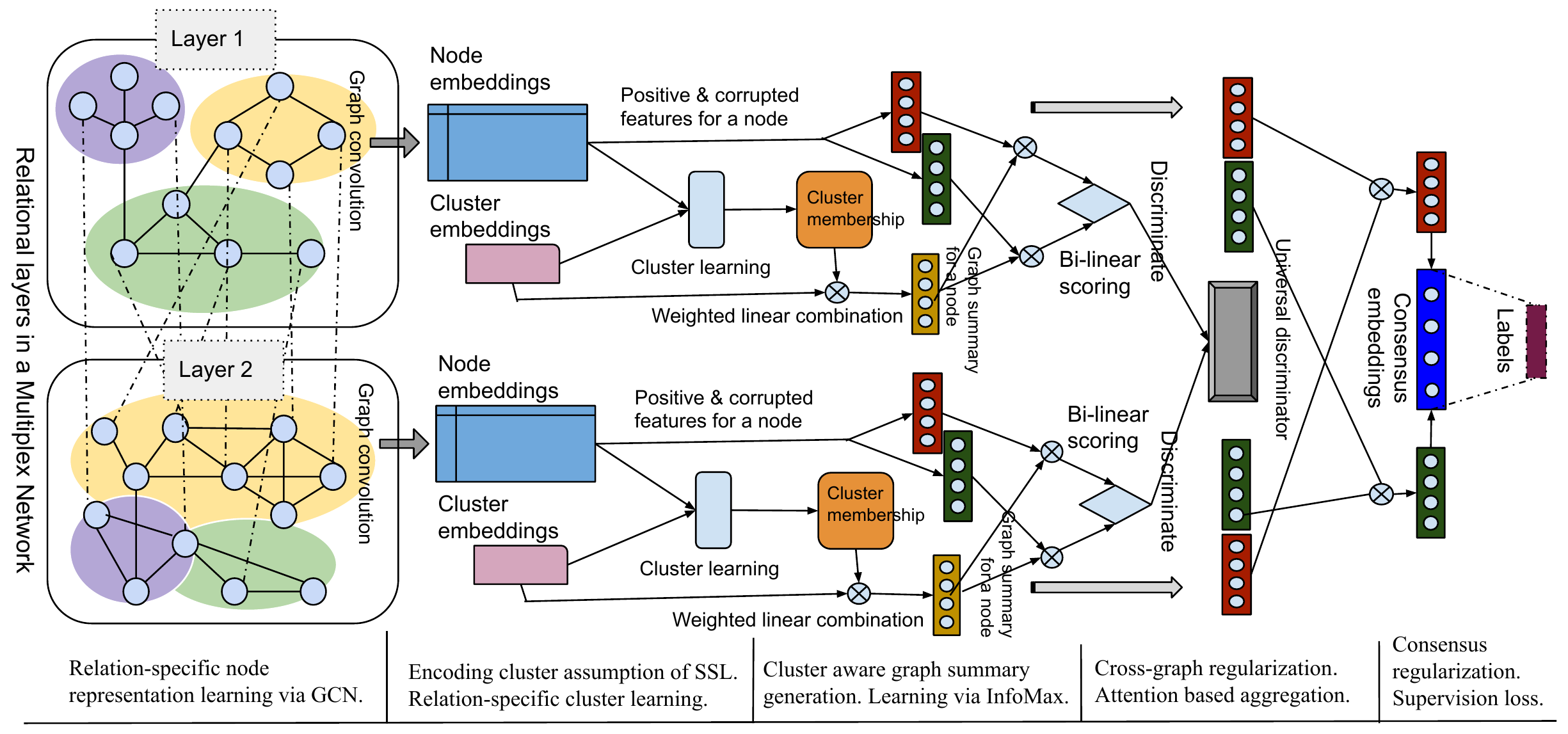}
\endminipage 
\caption{Semi-Supervised Deep Clustered Multiplex (SSDCM) with structure-aware graph summary generation
\\\scriptsize{Explanation of used color-codes. Green: True node embeddings, Red: Corrupted node embeddings, Blue: Final consensus node embeddings. Orange: Learned cluster membership for nodes, Purple: Label information. The color-coded grouping of nodes in relational layers of the example multiplex network denotes different clusters.}
}
\label{figure:architecture}
\end{figure*}

In this section, we explain step-by-step our proposed approach to learning node representations for multiplex networks. The proposed method Semi-Supervised Deep Clustered Multiplex (SSDCM) in Figure~\ref{figure:architecture} --- (i) learns relation-specific node representation that encodes both local and global information, (ii) enforces cross-edge based regularization to align all nodes connected across layers to lie on the same space, then (iii) learns a joint embedding across layers for all nodes through a consensus regularization and (iv) finally enables label predictions with this joint embedding. 

\subsection{Learning Node Representations} 
The first component of our model learns relation-specific ($\mathcal{R}$) local node representations, $U_r$. Then these learned node representations are made aware of their individual global context, $S_r$ which summarizes graph level information. We do this by maximizing the mutual information between them and this can be realized by minimizing a noise-contrastive estimation such as a binary cross-entropy loss as provided in the equation below,
\begin{align} \label{eqn:infomax}
\mathfrak{O}_{MI} &= \sum_{r \in \mathcal{R}} \sum_{i \in \mathcal{V}} \Big( \log(\mathfrak{D} (U_r^i, S_r^i))+ \sum_{j=1}^N \log(1-\mathfrak{D}(\tilde{U}_r^j, S_r^i)) \Big)
\end{align}
where  $\mathfrak{D}:\mathbb{R}^{2d} \mapsto \mathbb{R}$ is a discriminator function that assigns a probability score to a pair of local-global node representations using a bi-linear scoring matrix $B \in \mathbb{R}^{d \times d}$, i.e., {\color{m_blue}$ \mathfrak{D}(U_r^i, S_r^i)= \sigma({U^{i}_{r}}^{T} B S_r^i)$}, $\sigma$ being the sigmoid non-linearity. Similar to \cite{park2020unsupervised}, we learn this discriminator universally, i.e., the weight is shared across all layers with an intention to capture local-global representation correlations across the relations. The discriminator gives a local-global summary a higher value if the local summary is highly relevant to the global summary and, in contrast, assigns a lower score if the local summary is {\color{m_blue}less} relevant or irrelevant to the global summary. For every node $i$ in each relation $r \in \mathcal{R}$, $N$ negative local summaries are paired with {\color{m_blue}that node's} contextual global summary to train the discriminator $\mathfrak{D}$. Following \cite{velivckovic2018deep}, we create corrupted local patches $\tilde{U}_r^j$ for each relation $r$ by row-shuffling the node features $X$ and passing it through the same local structure encoder.

Having explained the overall structure of our InfoMax objective, we get into the details of how to learn (a) local node representations, (b) global node representations, and (c) the clustering strategy that provides
the global context for nodes.

\subsubsection{Local Node Representations}
For each relation $r \in \mathcal{R}$, we obtain an $M$-hop local node representations {\color{m_blue}$U_r \in \mathbb{R}^{|\mathcal{V}|\times d}$} with a Graph Convolutional Neural Network (GCN) encoder $\mathfrak{E_r}$. GCNs obtain an $M$-hop local representation by recursively propagating aggregated neighborhood information. 
Let $\tilde{A}_{(r,r)} = A_{(r,r)} + \epsilon I_{|\mathcal{V}|}$ be the intra-layer adjacency matrix for relation $r$ with added $\epsilon$-weighted self-loops (similar to a Random Walk with Restart (RWR) probability kernel). Here, we use the normalized adjacency matrix, $\hat{A}_{(r,r)} = (\tilde{D}^{-\frac{1}{2}}_{(r,r)}\tilde{A}_{(r,r)}\tilde{D}^{-\frac{1}{2}}_{(r,r)})$, as the GCN's neighborhood aggregation kernel, where $\tilde{D}_{(r,r)}$ is the diagonal degree matrix of $\tilde{A}_{(r,r)}$. An $M$-hop node embedding is obtained by stacking $M$-layers of GCNs as in Eqn: \ref{eqn:gcn}. The input to the $m$\textsuperscript{th} GCN layer is the output of the $(m-1)$\textsuperscript{th} GCN layer, $X_r^{m-1}$, with the original node features $X$ fed in as input to the first layer.
\begin{align} \label{eqn:gcn}
X_r^0 &= X \nonumber\\
X_r^m &= \textit{PReLU}(\hat{A}_{(r,r)}X_r^{m-1}W_r^m)\\
U_r &= X_r^M \nonumber
\end{align}
where $W_r^m$ is learnable weight matrix for the $m$\textsuperscript{th} GCN layer corresponding to the $r$\textsuperscript{th} multiplex layer. Assuming all GCN layers' outputs to be of the same dimension $d$, we have $X_r^m \in \mathbb{R}^{|\mathcal{V}|\times d}$ and $W_r^m \in \mathbb{R}^{d\times d}$, except for the first GCN layer, whose weights are $W_r^0 \in \mathbb{R}^{|\mathcal{F}|\times d}$. {\color{m_blue}The final $M$-hop GCN representation for each relation, $r$ is treated as that relation's local node embedding, $U_r = X_r^M$.}

\subsubsection{Contextual Global Node Representations} 
We learn a contextualized global summary representation, $S_r^i$ for each node $i$, and for every relation $r \in \mathcal{R}$. In this work, we first capture a global graph level summary by learning $K$ clusters in each relation. Then we leverage these learned clusters to provide a contextual global node summary for all the nodes based on the {\color{m_blue}learned node-cluster associations.} We explain the steps in a top-down manner. We first explain how we obtain a contextualized global graph representation given clustering information, and in the following subsection, we explain how to obtain the clusters.

Across all multiplex layers, we learn $K$ clusters in each relation $r \in \mathcal{R}$. We encode the clustering information with relation-specific $K$ cluster embeddings, $C_r = \{C_r^1, C_r^2, ..., C_r^K\}$ with $C_r^k \in \mathbb{R}^{1 \times d}$  and node-cluster assignment matrix, $H_r \in \mathbb{R}^{|\mathcal{V}|\times K}$. 
Given the learned relation-wise clustering information $(C_r, H_r)$ and local node representations, $U_r$, we compute the contextual global node representation for a node $i$ as a linear combination of different cluster embeddings, $C_r^k$ weighted by that node's cluster association scores $H_r^i[k], \forall k \in [1, K]$ as mentioned in below.
\begin{eqnarray} \label{eqn:global}
S^i_r = \sum_{k=1}^K {H_r^i[k]C_r^k}
\end{eqnarray}

\subsubsection{Clustering} 
We now describe how to learn clusters that capture useful global information for a node across all relations. Specifically, we aim to capture globally relevant label information that can enrich local node representations for the semi-supervised node classification task when jointly optimized for the MI between them across relations. 
To achieve this, we adapt \cite{mitra2020unified}'s Non-Negative Matrix Factorization formulation to learn label-correlated clusters to a Neural Network setup as follows. 

\textbf{Cluster Embedding:} We randomly initialize the set of cluster embeddings $C_r$ for each relation $r$ and allow them to be updated based on the gradients from the model's loss.

\textbf{Cluster Assignment:} We obtain the node-cluster assignment matrix, $H_r$, by computing the inner-product between node embeddings and cluster embeddings. We then pass it through a softmax layer to obtain normalized probability scores of cluster-memberships for each node, see Eqn: \ref{eqn:clust-assign}.
\begin{equation} \label{eqn:clust-assign}
H_r^i[k] = {\color{m_blue}\textit{SoftMax}(U_r^{i}.{C_r^{k}}^T)}
\end{equation}

\textbf{Non-overlapping Clustering Constraint:} To enforce hard cluster membership assignments, we regularize the cluster assignment $H_r$ with block-diagonal constraints. Specifically, we ensure the block size to be one, and the resulting orthogonality constraint enforces less overlap in node assignments between every pair of clusters. This constraint is expressed as a loss function below. 
\begin{equation} \label{eqn:clust-orth}
    \mathfrak{O}_{Orthogonal} = {\lVert H_r^TH_r-I_{K} \rVert}_F^2
\end{equation}

\textbf{Global Label-homogeneous Clustering Constraint:} 
To capture globally relevant information for each relational graph, we group nodes based on an aspect --- enforcing homogeneity within clusters. Precisely, we capture global label-correlation information with a similarity kernel, $\mathcal{S} \in \mathbb{R}^{|\mathcal{V}| \times |\mathcal{V}|}$ and cluster nodes according to it. The label similarity kernel is defined between the labeled nodes {\color{m_blue}$\mathcal{L}$ as $\mathcal{S}=Y_{[\mathcal{L}]}.{Y_{[\mathcal{L}]}}^T $.} We use a masking strategy to consider only the label information of training nodes for enforcing this. 

We now use a Laplacian regularizer to enforce smoothness on the cluster-assignments according to the label-similarity kernel $\mathcal{S}$ as given in the equation below,
\begin{equation} \label{eqn:clust-LapSmooth}
    \mathfrak{O}_{Learn} = \textit{Tr}(H_r^T\Delta(\mathcal{S})H_r)
\end{equation}
where $\Delta(\mathcal{S})$ is the un-normalized Laplacian of the similarity kernel. The above Laplacian smoothing constraint enforces nodes with similar labels to lie in the same/similar clusters. 

 Note that since this is shared across relations, it enforces similar clustering to be learned across relations. The learned clusters can still vary based on the relation-specific node embeddings, thus capturing global shared context across diverse graph structures. More importantly, notice that the label-similarity kernel can connect nodes that may be far away by a distance longer than the local ($M$) multi-hop context considered and even can connect two nodes that are not reachable from each other.

In the entire pipeline, cluster learning is facilitated by the following loss function:
$\mathfrak{O}_{Clus} = (\mathfrak{O}_{Learn} + \mathfrak{O}_{Orthogonal})$

\subsection{Cross-relation Regularization}
Since each multiplex layer encodes a different relational aspect of nodes, it is not straightforward to treat the inter-layer edges the same way as intra-layer edges to aggregate information from cross-linked neighbors. Also, the representation for nodes in different layers lies in different spaces and is not compatible with a naive aggregation of information via cross-edges.

\par Previous works, incorporated inter-layer (cross-graph) edge information into the learning procedure by adopting either cross-graph node embedding regularization \cite{li2018multi, ni2018co} or clustering techniques \cite{luo2020deep}. Since, in our case, we have the same clustering constraints enforced across layers, we opt to regularize embeddings of nodes connected by cross edges to lie in the same space.
{\color{m_blue}\begin{align} \label{eqn:14_1}
\mathfrak{O}_{Cross} = \sum_{r,s \in \mathcal{R}}
|| O_{(r,s)}U_r - A_{(r,s)}U_s||_F^2 
\end{align}}
where $O_{(r,s)} \in \mathbb{R}^{|\mathcal{V}| \times |\mathcal{V}|}$ is a binary diagonal matrix, with $O_{(r,s)}^{i,i}=0$ if the corresponding $A_{(r,s)}^i$ row-wise entries for node $i$ are all-zero and $O_{(r,s)}^{i,i} = 1$ otherwise if the bipartite association exists.This regularization aligns the representations of nodes that are connected by cross-edges to lie on the same space and be closer to each other.

\subsection{Joint embedding with Consensus Regularization}
Having obtained rich node representations that incorporated local, global and cross-layer structural information at every relational layer, we need a mechanism for aggregation of nodes' different relation-specific representations into a joint embedding. Since different layers may contribute differently to the end task, we use an attention mechanism to aggregate information across relations as, 
\begin{equation} \label{eqn:13}
\mathcal{J}_{r}^i = \frac{\exp(L_r \cdot U_r^i)}{\sum_{r' \in \mathcal{R}}\exp(L_{r'} \cdot U_r^i)}, \quad \quad U^i = \sum_{r \in \mathcal{R}} \mathcal{J}_{r}^iU_r^i
\end{equation} 
where $L_r$ is the layer-specific embedding and $\mathcal{J}_{r}^i$ is the importance of layer $r$ for node $i$. The importance score is computed by measuring the dot product similarity between the relational node embedding $U_r^i$ and learned layer embedding $L_r$. 
 
\par Additionally, to obtain a consensus embedding \cite{park2020unsupervised}, we leverage the corrupted node representations $\tilde{U_r} : r \in \mathcal{R}$ that we computed for InfoMax optimization in Eqn: \ref{eqn:infomax}. 
A consensus node embedding $Z \in \mathbb{R}^{|\mathcal{V}| \times d}$ is learned with a regularization strategy that minimizes the dis-agreement between $Z$ and attention-weighted aggregated true node representations $U$, while maximizing the dis-agreement between combined corrupted node representations $\tilde{U}$ (re-using the same attention weights). The final consensus node embedding $Z$ is generated as,
\begin{equation} \label{eqn:14}
\mathfrak{O}_{Cons} =  \mkern9mu {\lVert Z-U \rVert}_F^2 - {\lVert Z-\tilde{U} \rVert}_F^2
\end{equation}

\subsection{Semi-Supervised Deep Multiplex Clustered InfoMax}
We predict labels $\hat{Y}$ for nodes using their consensus embeddings $Z$. We project $Z$ into the label space  using weights $W_Y \in \mathbb{R}^{d \times |\mathcal{Q}|}$ and normalize it with $\sigma$, a \textit{softmax} or \textit{sigmoid} activation function for multi-class and multi-label tasks respectively. The prediction function is learned by minimizing the following cross-entropy loss,
\begin{align} \label{eqn:15}
\mathfrak{O}_{Sup} &= - \frac{1}{|\mathcal{L}|} \sum_{i \in \mathcal{L}} \sum_{q \in \mathcal{Q}} Y_{iq} \ln \hat{Y}_{iq} \\
\hat Y &= \sigma(ZW_Y) \nonumber
\end{align}
Finally, the overall semi-supervised learning process to obtain rich node representations that capture local-global structures in a multiplex network is obtained by jointly optimizing the equation below that optimizes different necessary components. We leverage hyper-parameters $\alpha, \beta, \gamma, \zeta, \theta$ to fine-tune the contributions of different terms. 
\begin{equation} \label{eqn:16}
\mathfrak{O} = \alpha * \mathfrak{O}_{MI} + \beta * \mathfrak{O}_{Cross} + \gamma * \mathfrak{O}_{Cons} + \zeta * \mathfrak{O}_{Clus}  + \theta * \mathfrak{O}_{Sup}
\end{equation}
Empirically, we find that our objective function is not very sensitive to variation in $\alpha, \beta$ values. Therefore, we fix their values as $\alpha=1.0, \beta=0.001$. Finally, we only tuned variables $\gamma, \zeta, \theta$ in the above objective function to analyze the contributions of network, cluster, and label information. {\color{m_blue}We discuss this further in Table~\ref{table:hyperparam_range} of Appendix~\ref{subsubsection:baselines_detailed}.}

\section{Related works} \label{section:related_works}
Here, we discuss related representation learning literature focused on multiplex networks. 

\par \textbf{Network Representation Learning (NRL).}
Network Representation Learning (NRL) methods for multiplex networks use different learning paradigms such as matrix factorization \cite{li2018multi}; random-walk based objectives \cite{liu2017principled, zhang2018scalable} and graph neural network architectures \cite{ghorbani2019mgcn, luo2020deep, park2020unsupervised}.

NRL models for multiplex networks have aimed at capturing different aspects of this multi-layered data. Modeling multi-layer data might require one to capture local \cite{li2018multi, matsuno2018mell} and global network structures \cite{luo2020deep, park2020unsupervised} within each layer; leverage cross-layer edges \cite{li2018multi, ghorbani2019mgcn, ni2018co, matsuno2018mell} between layers; encode node features \cite{ghorbani2019mgcn, park2020unsupervised}; integrate information from multiple layers into a unified feature space \cite{liu2017principled, zhang2018scalable}; {\color{m_blue}Optimize for single objective \cite{schlichtkrull2018modeling, wang2019heterogeneous} or jointly optimize for different objectives at different layers \cite{ghorbani2019mgcn, luo2020deep}.}

\par \textbf{Global context-based NRL.} 
In general, random-walk-based methods and GCNs are limited to capturing k-hop local contexts of nodes only. While matrix factorization methods embed the entire graph, they are neither scalable nor powerful than the other two. Only a few studies capture the global structures into node embedding learning in both homogeneous and multiplex networks.

GUNets~\cite{gao2019graph} introduced pooling and unpooling operations on multi-graphs for homogeneous networks. Its prominence-based node pooling captures the global graph structures and local graph structures using learnable projection vectors and GCNs. Deep Multi-Graph Clustering (DMGC) \cite{luo2020deep} is the first NRL study to learn global structures for multilayer networks explicitly. It proposes an attentive unsupervised mechanism to encode the cluster structures into multi-graphs based on a similarity-based cluster kernel. 

\par \textbf{InfoMax based NRL.} {\it Deep Multiplex Graph Infomax (DMGI)}\cite{park2020unsupervised} employs the InfoMax principle for multiplex networks. It jointly maximizes the MI between the local and global graph patches across the layers of a multiplex graph. It does so by learning a universal discriminator that discriminates positive and negative patch pairs across the relational layers. Simultaneously, it employs a regularization strategy that attentively aggregates the learned relation-specific node representations by reusing negative node representations used for learning the discriminator weights. HDGI~\cite{ren2020hdgi} is a work similar to DMGI, aimed at heterogeneous networks. It adopts a semantic attention mechanism to aggregate metapath influenced node embeddings and discriminator-based learning strategy. 

\par \textbf{Semi-Supervised Learning (SSL).} 
State-of-the-art methods MGCN \cite{ghorbani2019mgcn}, DMGI \cite{park2020unsupervised} are examples of SSL frameworks for multilayer/ multiplex networks. MGCN proposes a layered graph convolution neural (GCN) architecture to preserve the within layer and cross-layer network structures by leveraging a cross-entropy loss function in each network layer. DMGI -- though originally proposed as an unsupervised method, inculcates a semi-supervised variant that explicitly guides the learning of layer attention weights. We have HAN~\cite{wang2019heterogeneous} and RGCN~\cite{schlichtkrull2018modeling} from the domains of heterogeneous and knowledge graphs, respectively. HAN proposes a GNN architecture based on hierarchical node-level and metapath-level attention mechanisms. In contrast, RGCN proposes a graph convolution-based message passing framework facilitating effective weight sharing to avoid overfitting on rare relations.

\par In Table~\ref{table:baseline_compare}, we summarize competing methods in terms of important aspects of a multi-graph that they are designed to capture. These methods either lack strategies for 1) capturing \emph{global structural information:} or 2) \emph{aggregating} node information  across different counterparts of the same node from different layers, 3) capturing useful \emph{structures}. Even if there are global NRL methods like DMGI, DMGC, GUNets --- they either use a naive mean-pooling approach for acquiring global graph representations or unsupervised clustering criteria/ importance pooling strategy to capture global graph structures that might not be useful given the end task is concerned. Our framework, SSDCM,  differs in that we build upon a semi-supervised structure-aware version of \emph{InfoMax} --- which is first-of-its-kind to the best of our knowledge. Our objective is to learn global-structure enhanced node representations suitable for node-wise tasks capturing all aspects of multiplex graphs.

\begin{table}[h] \centering
	\setlength\extrarowheight{2pt}
	\begin{adjustbox}{max width=0.80\textwidth}
		\begin{tabular}{l|l|l|l|l|l|l|l|l|l|}
			\cline{2-9}
			\multirow{2}{*}{}                            & \multirow{2}{*}{Methods} &
			\multicolumn{7}{c|}{Comparison}  
			\\ \cline{3-9} 
			&                & \rotatebox{90}{\textbf{\textsc{DMGC}}~\cite{luo2020deep}\phantom{.}}   & \rotatebox{90}{\textbf{\textsc{DMGI}}~\cite{park2020unsupervised}\phantom{.}}  & \rotatebox{90}{\textbf{\textsc{HAN}}~\cite{wang2019heterogeneous}\phantom{.}}  & \rotatebox{90}{\textbf{\textsc{MGCN}}~\cite{ghorbani2019mgcn}\phantom{.}}  & \rotatebox{90}{\textbf{\textsc{RGCN}}~\cite{schlichtkrull2018modeling}\phantom{.}}  & \rotatebox{90}{\textbf{\textsc{GUNets}}~\cite{gao2019graph}\phantom{.}} & \rotatebox{90}{\textbf{\textsc{SSDCM}}\phantom{.}}    \\ \hline
			\multicolumn{1}{|l|}{\multirow{6}{*}{\rotatebox{90}{Properties}}} & attributes   &  &\checkmark   &\checkmark   &\checkmark   &    &\checkmark   & \checkmark  \\ \cline{2-9} 
			\multicolumn{1}{|l|}{}                       & within-network &\checkmark   &\checkmark  &\checkmark  &\checkmark   &\checkmark  &\checkmark   &\checkmark  \\ 
			\cline{2-9} 
			\multicolumn{1}{|l|}{}                       &  cross-network &\checkmark  &  &\textbf{--}   &\checkmark  &\textbf{--}   &\textbf{--}  &\checkmark    \\ 
			\cline{2-9} 
			\multicolumn{1}{|l|}{}                       &  labels &  &\checkmark  &\checkmark   &\checkmark   &\checkmark   & \checkmark     & \checkmark   \\
			\cline{2-9} 
			\multicolumn{1}{|l|}{}                       &  global structure  &\checkmark  &\checkmark  &   &   &   &\checkmark  & \checkmark   \\
			\cline{2-9} 
			\multicolumn{1}{|l|}{}                       &  aggregation   &  &\checkmark  &\checkmark   &\checkmark   &\checkmark   &    &\checkmark    
			\\\hline
			\multicolumn{9}{l}{* \scriptsize{Dash marks denote Not Applicable (NA).}}
		\end{tabular}
		\caption{Coverage of multiplex network features. {\scriptsize{{\color{m_blue}}}}} \label{table:baseline_compare}
	\end{adjustbox}
\end{table}
\vspace{-1em}
\begin{table}[h]
	\centering
	\begin{adjustbox}{max width=0.80\textwidth}
		\begin{tabular}{@{}lccccc@{}}
			\toprule
			\textbf{Dataset} & \textbf{Layers} & \textbf{Nodes} & \textbf{Edges}(Total) & \textbf{Features} & \textbf{Labels} \\ \midrule
			\textbf{ACM}~\cite{wang2019heterogeneous}              & 5               & 7427           & 24536689    & 767   & 5               \\
			\textbf{DBLP}~\cite{wang2019heterogeneous}             & 4               & 4057           & 17976710   & 8920    & 4               \\
			\textbf{SLAP}~\cite{zhang2018deep}             & 6               & 20419          & 8207130   & 2695     & 15              \\
			\textbf{IMDB-MC}~\cite{park2020unsupervised}          & 2               & 3550           & 80216   & 2000       & 3               \\
			\textbf{IMDB-ML}~\cite{pham2017column}          & 3               & 18352          & 2505797    & 1000    & 9               \\
			\textbf{FLICKR}~\cite{luo2020deep}           & 2               & 10364          & 506051    & --     & 7               \\
			\textbf{AMAZON}~\cite{park2020unsupervised}           & 3               & 17857          & 2194389 & 2395        & 5               \\ \bottomrule
		\end{tabular}
		\caption{Statistics of datasets \scriptsize{(Refer to Table~\ref{table:dataset}, Appendix~\ref{subsubsection:datasets_detailed} for details)}}\label{table:tiny_stats}
	\end{adjustbox}
\end{table}
\section{Experimental Setup} \label{section:experiments}
\textbf{Datasets.} \label{subsection:datasets} We evaluate our proposed algorithm SSDCM on a variety of datasets as mentioned in Table~\ref{table:tiny_stats}, from diverse domains, containing --- both multi-class and multi-label datasets, as well as, attributed and non-attributed datasets. 
Refer to Table~\ref{table:dataset}, Appendix~\ref{subsubsection:datasets_detailed} for additional dataset details.
\\\textbf{Baselines.} \label{subsection:baselines}
We chose State-Of-The-Art (SOTA) competing methods applicable to a diverse range of multi-graph settings. The compared methods can be roughly categorized into the following classes: multi-layered network-based embedding approaches ---  DMGC, MGCN; multiplex network embedding --- DMGI; heterogeneous network embedding --- HAN; multi-relational network embedding --- RGCN; pooling method in multi-graph setting --- GUNets. 
\\\textbf{Evaluation Strategy.} \label{subsection:experiment_setup}
We use a random sampling strategy to split the nodes into train, validation, and test set. We choose one-third of the labeled examples as train nodes. We keep the validation set size as half of the train set size. Thus, half of the total nodes are kept for evaluation purposes as test-set. Our experimental setup is summarized in Table~\ref{table:hyperparam_range}, Appendix~\ref{subsubsection:baselines_detailed}. For the methods applicable to non-attributed graphs, namely, RGCN and DMGC -- we implement attributed versions. For RGCN, we customized the relational GCN to take node features as input. For DMGC that uses relation-specific autoencoders to reconstruct the layer adjacencies, we input another array of feature-specific autoencoders. The feature-based autoencoders jointly learn a common hidden node representation along with the relational autoencoders and reconstruct layers' node features. To set up attributed NRL methods for FLICKR, we leverage the layer adjacencies as node features. We simply obtain the average of layer node embeddings as final representations for the methods with no specific node embedding aggregation strategy. 
\\We provide additional details of our analysis to facilitate replicability of results in Appendix~\ref{subsubsection:reproducibility}. We also provide our code \footnote{\url{https://github.com/anasuamitra/ssdcm}}.

\section{Results} \label{subsection:results}
We demonstrate the effectiveness of the proposed framework on four tasks, namely, node classification, node clustering, visualization, and similarity search. The details of the task-specific experiment setup, along with insights on results, are discussed below.
\subsection{Node Classification} \label{subsubsection:node_classification} 
For semi-supervised methods, we use the predicted labels directly to compute the node classification scores based on ground-truth. We train a logistic regression classifier on the learned node embeddings of the training data for unsupervised methods and report the performance of the predictor on the test node embeddings averaged over twenty runs. We report the test-set performance that corresponds to the best validation-set performance for a fair comparison. Micro-F1 and Macro-F1 scores are reported as node classification results in Tables~[\ref{table:micro_f1}, \ref{table:macro_f1}] respectively.
\begin{table}[h]\centering 
\begin{adjustbox}{max width=0.75\textwidth}
\begin{tabular}{@{}c|ccccccc@{}}
\toprule
\textbf{Micro-F1}   & \textbf{ACM}    & \textbf{DBLP}   & \textbf{SLAP}   & \textbf{FLICKR} & \textbf{AMAZON} & \textbf{IMDB-MC} & \textbf{IMDB-ML} \\ \midrule
\textbf{DMGC}       & 42.822          & 84.684          & 29.819          & 50.308          & 69.716          & 56.278           & 44.765           \\
\textbf{RGCN}       & 39.118          & 83.514          & 26.914          & 82.69           & 72.957          & 62.542           & 49.802           \\
\textbf{GUNets} & 46.428          & 87.124          & {\ul 32.985} & 87.607          & 77.177          & 52.508           & 43.988           \\
\textbf{MGCN}       & 52.458          & 87.003          & 29.563          & {\ul 91.307}    & 84.083          & 63.384           & 48.059           \\
\textbf{HAN}        & 77.441          & 85.989          & 30.976          & 89.478          & 83.77           & 62.353           & 47.117           \\
\textbf{DMGI}       & {\ul 81.205}    & {\ul 89.43}     & 30.03           & 91.225          & {\ul 89.422}    & {\ul 65.21}      & {\ul 53.413}     \\ \hline
\textbf{SSDCM}      & \textbf{88.324} & \textbf{94.988} & \textbf{33.597}    & \textbf{96.261} & \textbf{92.195} & \textbf{67.796}  & \textbf{54.055}  \\ \bottomrule
\end{tabular}
\end{adjustbox} \caption{Node classification results: Micro-F1 scores (\%)}\label{table:micro_f1}
\end{table}
\begin{table}[h]\centering 
\begin{adjustbox}{max width=0.75\textwidth}
\begin{tabular}{@{}c|ccccccc@{}}
\toprule
\textbf{Macro-F1}   & \textbf{ACM}    & \textbf{DBLP}   & \textbf{SLAP}   & \textbf{FLICKR} & \textbf{AMAZON} & \textbf{IMDB-MC} & \textbf{IMDB-ML} \\ \midrule
\textbf{DMGC}       & 39.679          & 83.279          & 21.581          & 46.122          & 64.013          & 54.699           & 29.122           \\
\textbf{RGCN}       & 38.665          & 82.86           & 24.119          & 81.47           & 68.323          & 62.17            & 45.285           \\
\textbf{GUNets} & 41.433          & 86.426          & 18.807          & 85.708          & 74.332          & 51.039           & 27.591           \\
\textbf{MGCN}       & 46.853          & 85.462          & {\ul 25.717}    & 91.07           & 82.349          & 62.876           & 38.821           \\
\textbf{HAN}        & 78.009          & 85.154          & 25.413          & 89.174          & 82.344          & 61.891           & 35.181           \\
\textbf{DMGI}       & {\ul 80.802}    & {\ul 88.828}    & 24.854          & {\ul 91.928}    & {\ul 88.114}    & {\ul 65.066}     & {\ul 48.122}     \\ \hline
\textbf{SSDCM}      & \textbf{88.571} & \textbf{94.681} & \textbf{28.072} & \textbf{96.147}  & \textbf{91.973} & \textbf{67.803}  & \textbf{51.756}  \\ \bottomrule
\end{tabular}
\end{adjustbox} \caption{Node classification results: Macro-F1 scores (\%)}\label{table:macro_f1}
\end{table}
From Tables~[\ref{table:micro_f1}, \ref{table:macro_f1}], it is clear SSDCM is the best performing model on all the datasets, by a significant margin. In comparison, DMGI gives the second-best performance on most datasets except on SLAP and FLICKR (for Micro-F1 scores).

\subsection{Node Clustering} \label{subsubsection:node_clustering}
\begin{table}[hb] \centering
\begin{adjustbox}{max width=0.750\textwidth}
\begin{tabular}{@{}c|ccccccc@{}}
\toprule
\textbf{NMI-N}      & \textbf{ACM}   & \textbf{DBLP}  & \textbf{SLAP}  & \textbf{FLICKR} & \textbf{AMAZON} & \textbf{IMDB-MC} & \textbf{IMDB-ML} \\ \midrule
\textbf{DMGC}       & 0.421          & 0.532          & 0.245          & 0.488           & 0.468           & 0.185            & 0.076            \\
\textbf{RGCN}       & 0.324          & 0.559          & 0.24           & 0.715           & 0.405           & 0.193            & \textbf{0.102}   \\
\textbf{GUNets} & 0.65           & {\ul 0.742}    & 0.251          & 0.758           & 0.519           & 0.108            & 0.036            \\
\textbf{MGCN}       & 0.41           & 0.738          & {\ul 0.278}    & {\ul 0.76}      & 0.528           & {\ul 0.195}      & 0.033            \\
\textbf{HAN}        & {\ul 0.939}    & 0.66           & {\ul 0.278}    & 0.639           & 0.519           & 0.178            & 0.055            \\
\textbf{DMGI}       & 0.837          & 0.682          & 0.275          & 0.644           & {\ul 0.568}     & 0.194            & 0.056            \\ \hline
\textbf{SSDCM}      & \textbf{0.947} & \textbf{0.819} & \textbf{0.284} & \textbf{0.822}  & \textbf{0.635}  & \textbf{0.223}   & {\ul 0.085}      \\ \bottomrule
\end{tabular}
\end{adjustbox}  \caption{Node clustering results: NMI scores}
\label{table:node_clustering}
\end{table} 
We only cluster the test nodes to evaluate performance on the node clustering task. We give the test node embeddings to the clustering algorithm as input to predict the clusters. We run each experiment ten times and report the average scores in Table~\ref{table:node_clustering}. K-Means and Fuzzy C-Means algorithms are used to predict clusters in multi-class and multi-label data, respectively. For multi-label data, we take the top $q$ number of predicted clusters, where $q$ is the number of classes that a node is associated with, to compare against the set of ground-truth clusters. We evaluate the obtained clusters against ground truth classes and report the Normalized Mutual Information (NMI) \cite{manning2008introduction} scores. We use Overlapping NMI (ONMI) \cite{lancichinetti2009detecting} for overlapping clusters to evaluate the multi-label datasets. Here we consider two kinds of clustering to demonstrate the effectiveness of our method. One is node clustering through clustering algorithms that takes final node embeddings as input. We refer to this clustering score as \emph{NMI-N}. Another is directly predicting clusters from the cluster membership matrices learned during the optimization process and comparing it to the ground-truth to evaluate the clustering performance. The latter score, referred to as \emph{NMI-C}, is only applicable to SSDCM and DMGC. From Table~\ref{table:node_clustering}, we can see that except on IMDB-ML, SSDCM outperforms all the competing methods on the clustering task. It beats the second-best performing model by $0.037$, across all datasets on average --- a significant improvement. 
\subsection{t-SNE Visualizations}\label{subsection:visualizations}
\begin{figure*}[ht] \scriptsize \centering
\minipage{0.165\textwidth}
  \includegraphics[width=\linewidth, keepaspectratio]{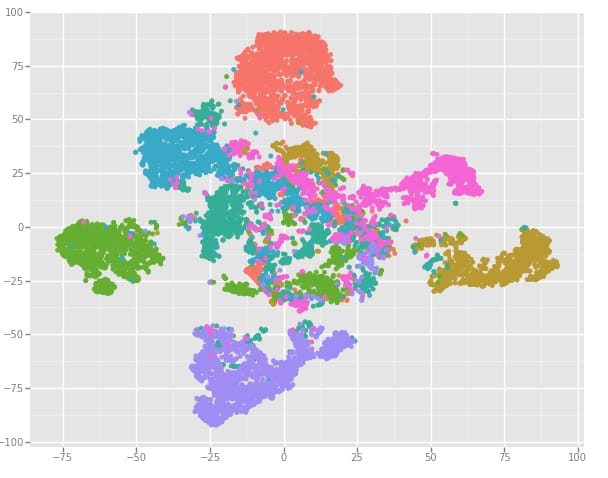}
\endminipage
\minipage{0.165\textwidth}
  \includegraphics[width=\linewidth, keepaspectratio]{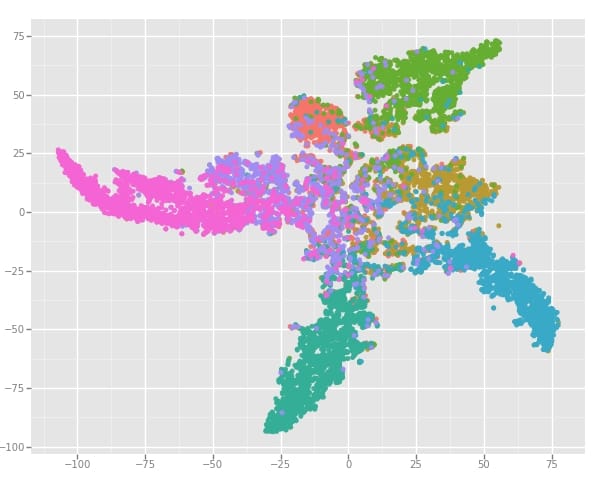}
\endminipage
\minipage{0.165\textwidth}
  \includegraphics[width=\linewidth, keepaspectratio]{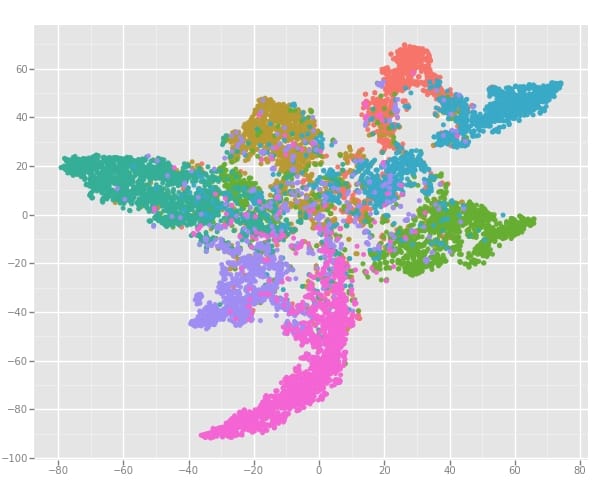}
\endminipage
\minipage{0.165\textwidth}
  \includegraphics[width=\linewidth, keepaspectratio]{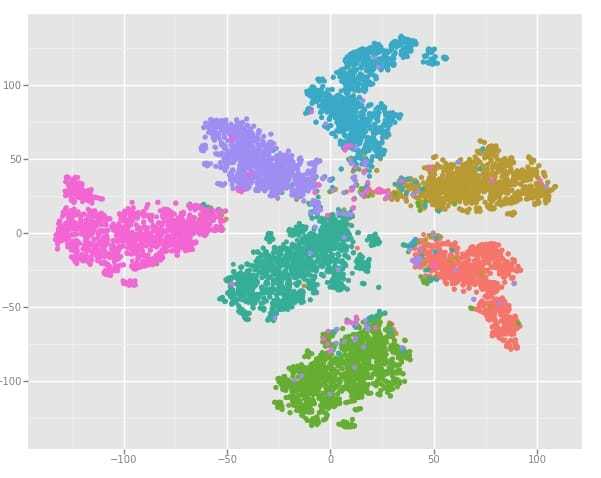}
\endminipage
\minipage{0.165\textwidth}
  \includegraphics[width=\linewidth, keepaspectratio]{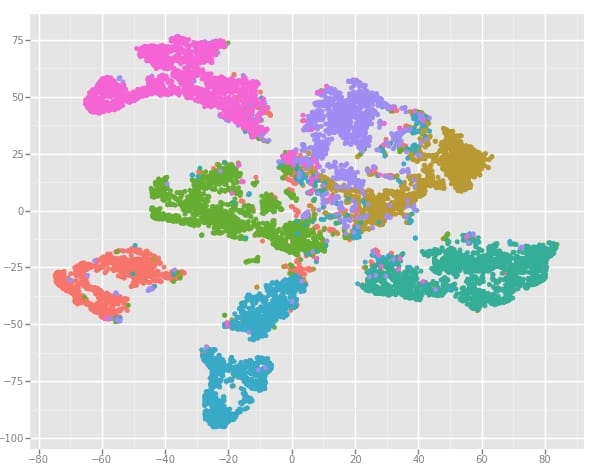}
\endminipage
\minipage{0.165\textwidth}
  \includegraphics[width=\linewidth, keepaspectratio]{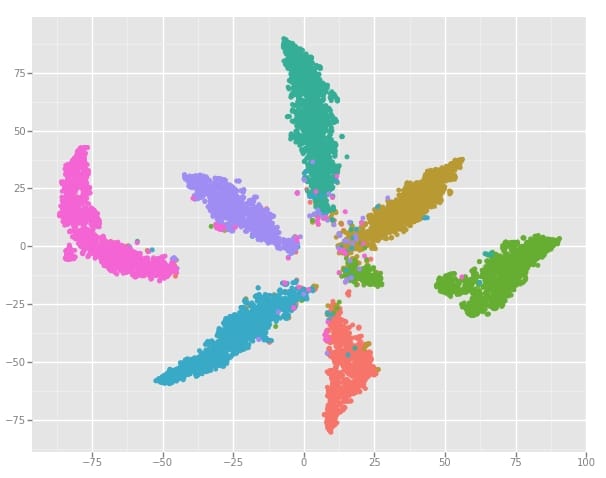}
\endminipage 
\\ 
\minipage{0.165\textwidth}
  \includegraphics[width=\linewidth, keepaspectratio]{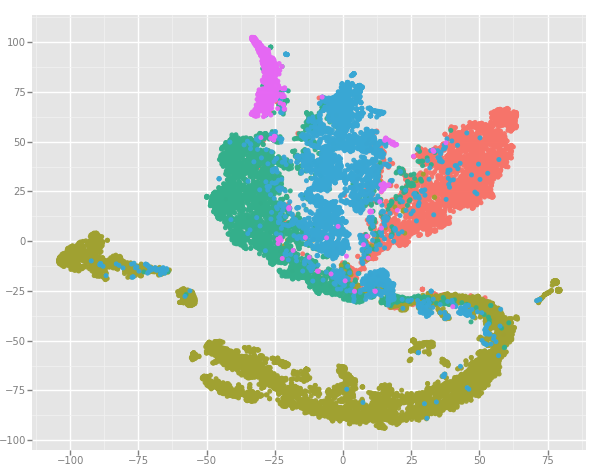}
  \begin{center} \subcaption{\scriptsize RGCN} \end{center}
\endminipage
\minipage{0.165\textwidth}
  \includegraphics[width=\linewidth, keepaspectratio]{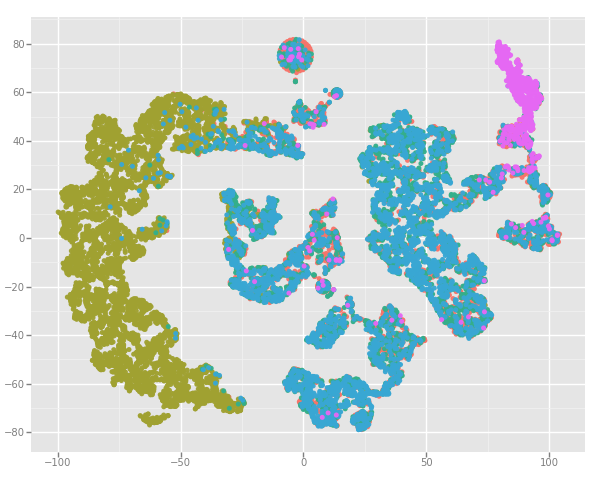}
  \begin{center} \subcaption{\scriptsize GUNets} \end{center}
\endminipage
\minipage{0.165\textwidth}
  \includegraphics[width=\linewidth, keepaspectratio]{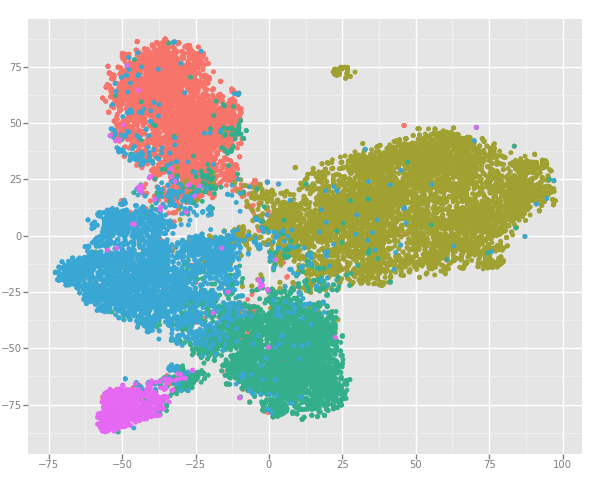}
  \begin{center} \subcaption{\scriptsize MGCN} \end{center}
\endminipage
\minipage{0.165\textwidth}
  \includegraphics[width=\linewidth, keepaspectratio]{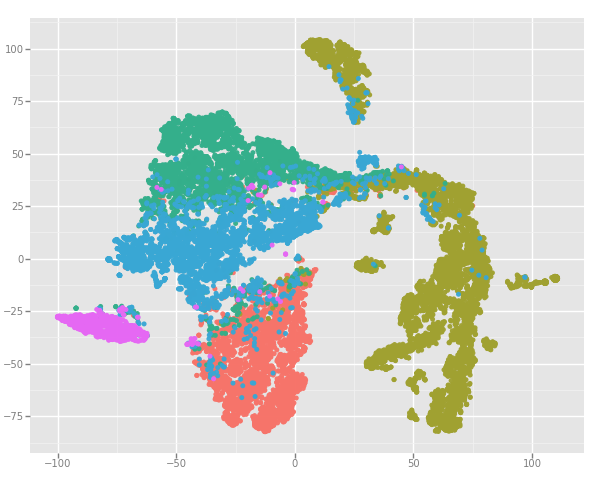}
  \begin{center} \subcaption{\scriptsize HAN} \end{center}
\endminipage
\minipage{0.165\textwidth}
  \includegraphics[width=\linewidth, keepaspectratio]{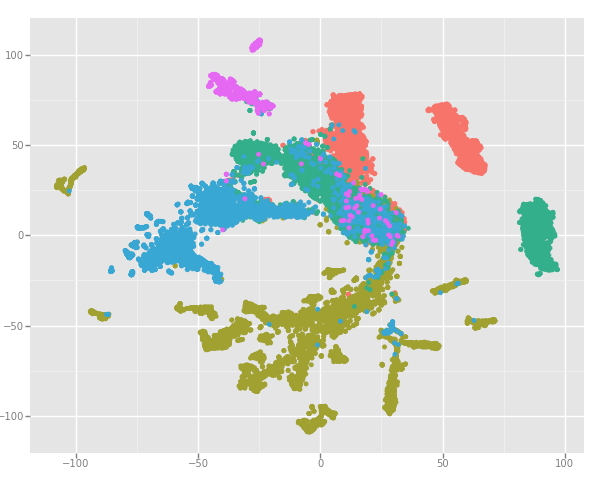}
  \begin{center} \subcaption{\scriptsize DMGI} \end{center}
\endminipage
\minipage{0.165\textwidth}
  \includegraphics[width=\linewidth, keepaspectratio]{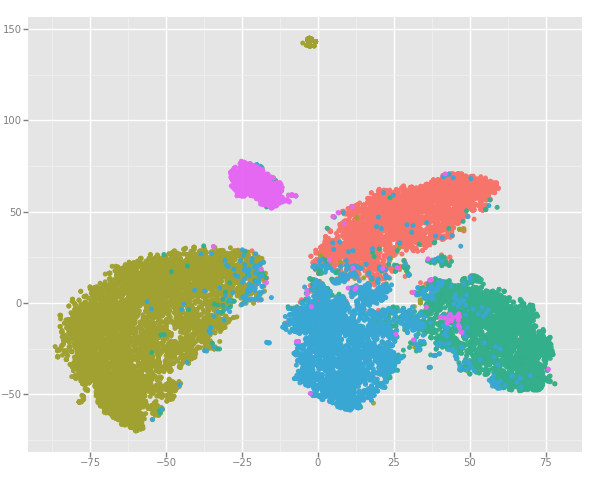}
  \begin{center} \subcaption{\scriptsize SSDCM} \end{center}
\endminipage \centering
\caption{t-SNE Visualization of node embeddings on FLICKR (top), AMAZON (bottom) for all the SSL methods}\label{figure:tsne}
Please refer to Section: \ref{subsection:baselines} for the candidate methods for which the t-SNE visualizations are plotted here. The color codes indicate functional classes (FLICKR: 7, AMAZON: 5).
\end{figure*}
We also visualize the superior clusterability of SSDCM's learned node representations for the FLICKR and AMAZON dataset in Figure~\ref{figure:tsne} using t-Distributed Stochastic Neighbor Embedding (t-SNE) \cite{van2008visualizing} visualization.
The color code indicates functional classes for respective datasets. We choose the node embeddings that gave the best performance in node classification scores for all the competing methods. We can see that all the semi-supervised methods yield interpretable visualizations indicating clear inter-class separation. Among them, SSDCM obtains compact well-separated small clusters of the same class labels, which appear to be visually better-separated than the rest of the methods. We see similar trends in visualization for other datasets also (not shown here).
\subsection{Node Similarity Search} \label{subsubsection:similarity_search}
\begin{figure}[h]
\minipage{0.2475\textwidth}
  \includegraphics[width=\linewidth, keepaspectratio]{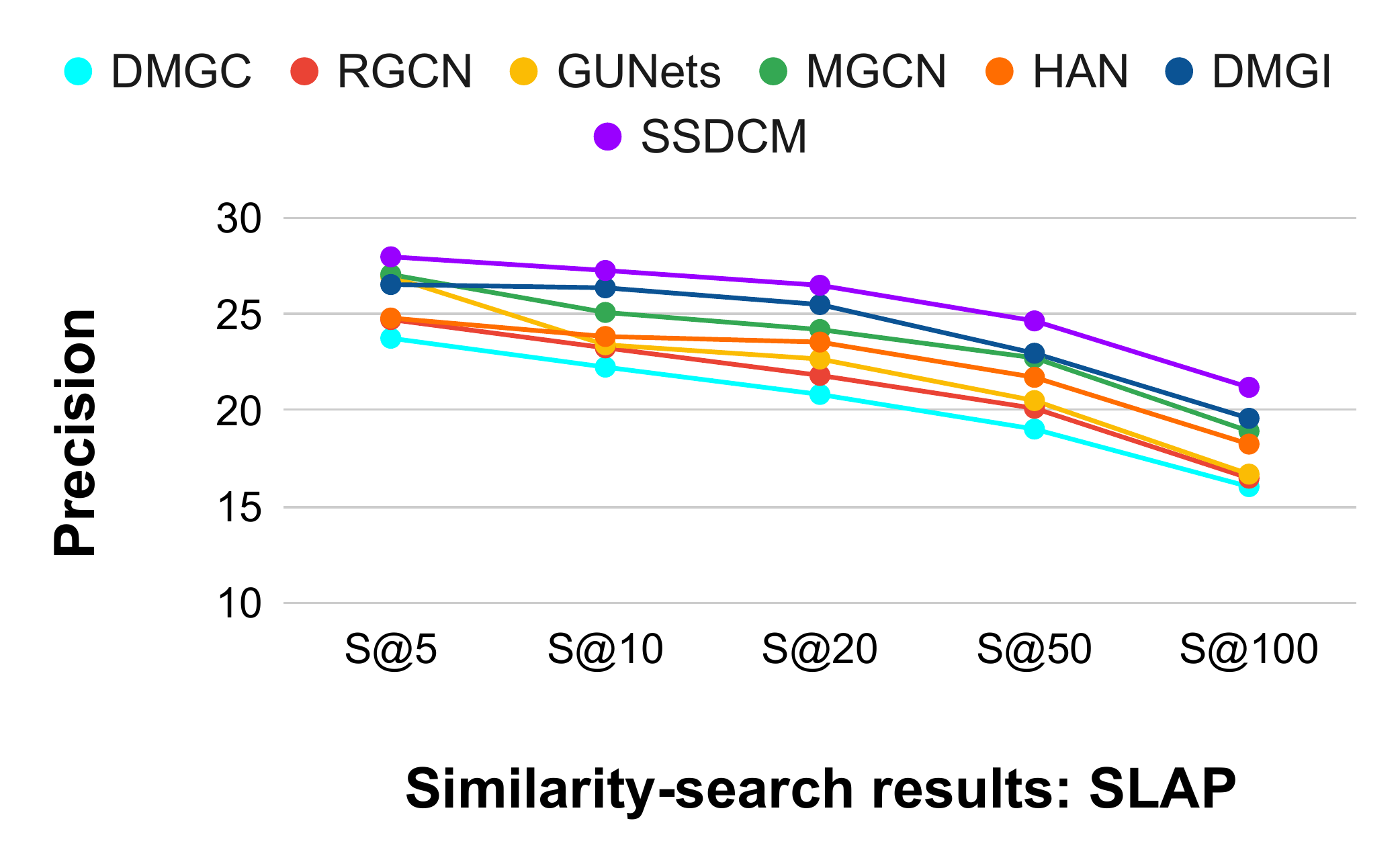}
\endminipage 
\minipage{0.2475\textwidth}
  \includegraphics[width=\linewidth, keepaspectratio]{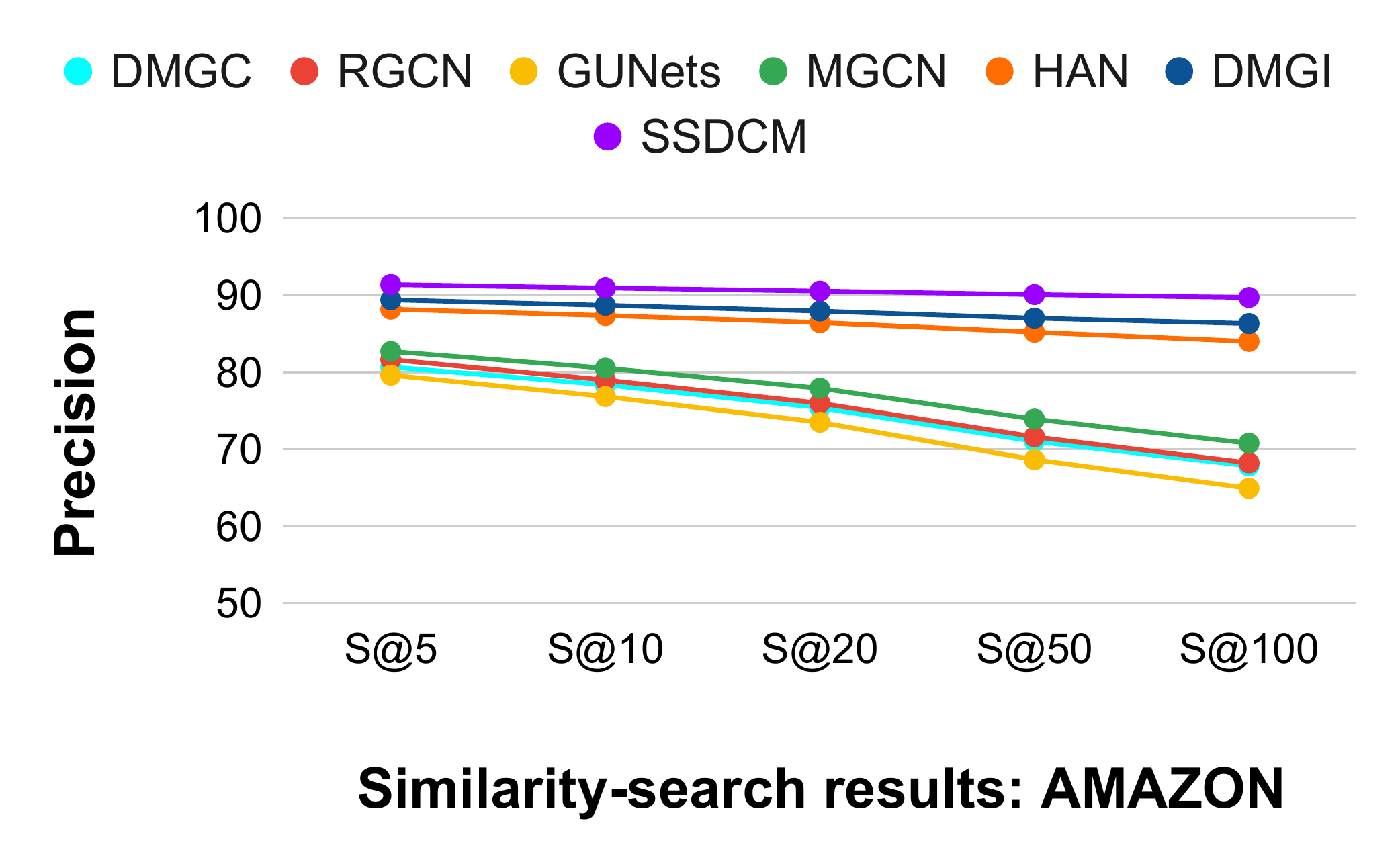}
\endminipage 
\minipage{0.2475\textwidth}
  \includegraphics[width=\linewidth, keepaspectratio]{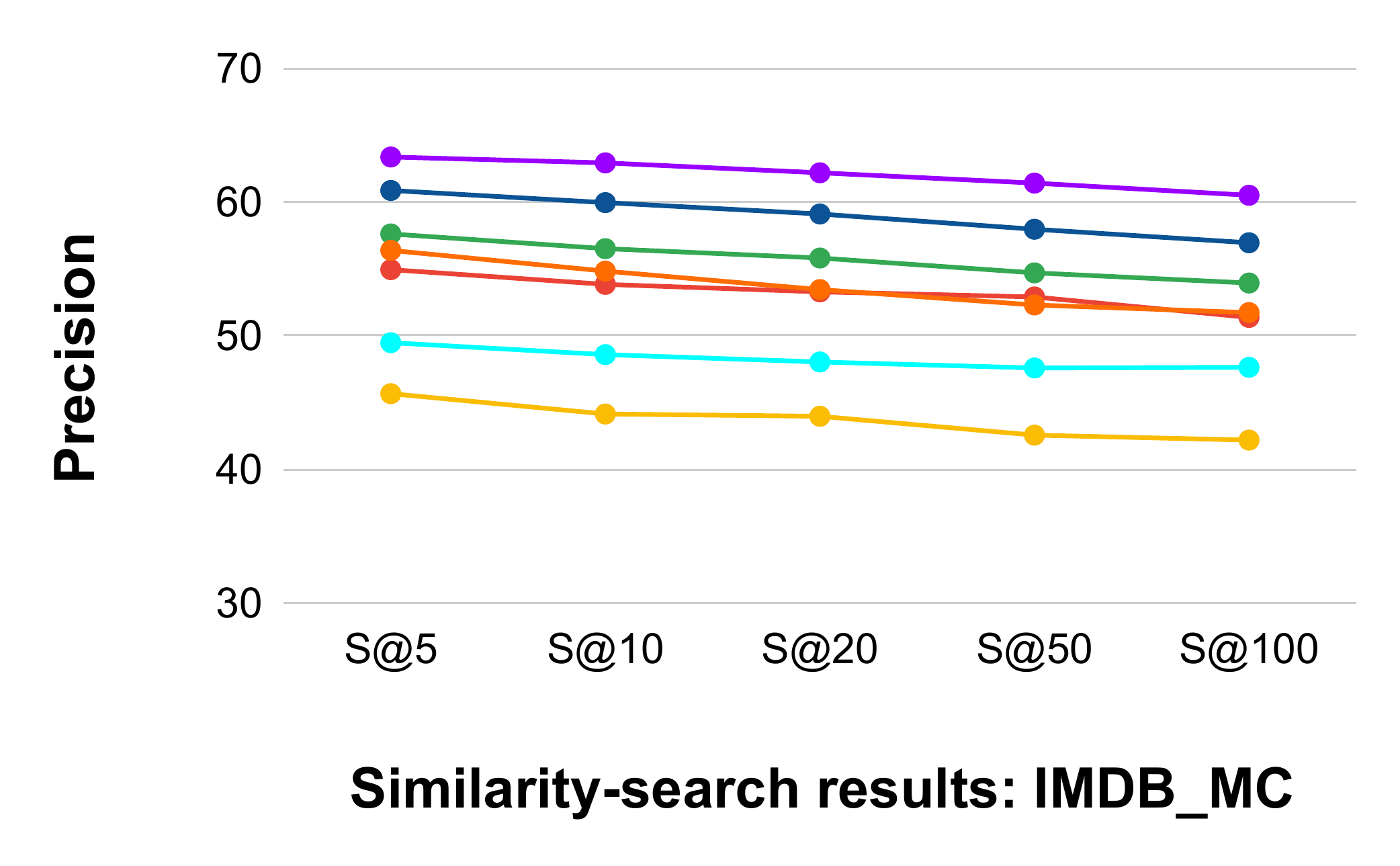}
\endminipage 
\minipage{0.2475\textwidth}
  \includegraphics[width=\linewidth, keepaspectratio]{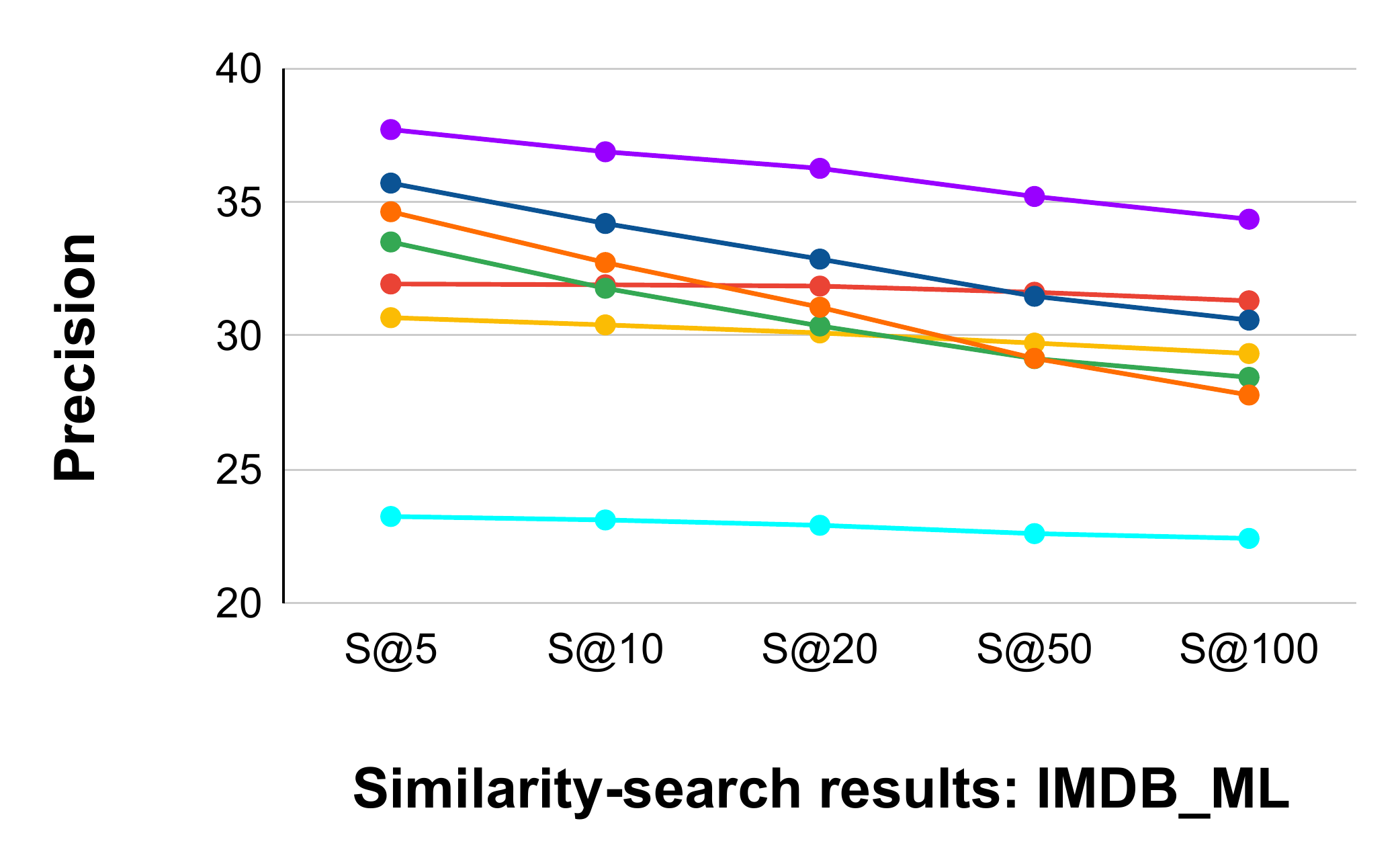}
\endminipage 
\caption{Comparing similarity search results}\label{figure:similarity_search}
\end{figure}
In a similar setup to  \cite{park2020unsupervised}, we calculate the cosine similarity scores of embeddings among all pairs of nodes. For a query node, the rest of the nodes are ranked based on the similarity scores. We then retrieve top $K=\{5, 10, 20, 50, 100\}$ nodes to determine the fraction of retrieved nodes with the same labels as the query node, averaged by $K$. For multi-label graphs, instead of exact label matching, we use the Jaccard similarity to determine the relevance of the query and target nodes' label set. We compute this similarity search score for all nodes as a query and report the average. The similarity search results get a significant boost under our framework since our encoding of the SSL clusters puts nodes with similar labels together in the same cluster. Whereas DMGC's clustering criterion, DMGI's global pooling, and GUNet's node importance based pooling criterion -- do not demonstrate a similar benefit. 
From Tables~[\ref{table:micro_f1}, \ref{table:macro_f1}], we see for SLAP and two versions of IMDB movie networks the classification score of the competing methods are close. But in similarity search, we can differentiate SSDCM as the best performing model among all. DMGI is the second-best performing model in node similarity search, similar to the node classification results. GUNets and DMGC are seen to perform worse than the rest. 
\section{Analysis} \label{section:case_studies}
Herein we conduct an array of drill down experiments to shed light on the key components of our proposed SSDCM framework. 

\subsection{Novelty of cluster-based graph summary} \label{subsection:cluster_pooling_novelty}
\begin{table}[h] \centering
\begin{adjustbox}{max width=0.750\textwidth}
\begin{tabular}{@{}l|ccccc@{}}
\toprule
Micro-F1 Scores                                                                    & IMDB\_MC        & ACM             & DBLP            & AMAZON          & FLICKR          \\ \midrule
SSDCM {[}global pool{]}                                                            & 65.942          & 84.176          & 91.592          & 90.62           & 92.698          \\
SSDCM {[}top-K pool{]}                                                             & 63.908          & 84.218          & 90.683          & 90.34           & 93.714          \\
SSDCM {[}SAG pool{]}                                                               & {\ul 66.574}    & 83.176          & {\ul 92.859}    & {\ul 90.878}    & 93.015          \\
SSDCM {[}ASAP pool{]}                                                              & 66.365          & {\ul 85.064}    & 91.782          & 90.844          & {\ul 94.689}    \\ \hline
\begin{tabular}[c]{@{}l@{}}SSDCM {[}cluster aware \\ graph summary{]}\end{tabular} & \textbf{67.796} & \textbf{88.324} & \textbf{94.988} & \textbf{92.195} & \textbf{96.261} \\ \bottomrule
\end{tabular}\end{adjustbox}  \caption{Novelty of cluster-based graph summary}
\label{table:cluster_pooling_novelty}
\end{table}
In Table~\ref{table:cluster_pooling_novelty}, we delve deeper into how good the cluster-aware graph summary representation (Equation~\ref{eqn:global}) is for the universal discriminator (Equation~\ref{eqn:infomax}). We consider alternative  SOTA pooling methods --- Top-K~\cite{gao2019graph}, SAG~\cite{zhang2020structure} and ASAP~\cite{ranjan2020asap} for generating graph summaries in the SSDCM framework. Top-K pool realizes node importance based pooling strategy via learning a projection vector. In comparison, the SAG pool improves upon the former by encoding structural information from graphs using GNNs. Adaptive Structure Aware Pooling (ASAP) is a new SOTA method that considers the cluster structures from graphs. It proposes a self-attentive GCN architecture \emph{Master2Token} to learn clusters and uses a cluster fitness-based scoring strategy to pool underlying graph structures in phases \footnote{We use the Pytorch Geometric \cite{Fey/Lenssen/2019} library for candidate pooling methods.} These pooling strategies generate a common graph summary for the whole graph, which is fed to the discriminator along with the node embeddings. On the contrary, our cluster-aware graph summary has a node's perspective, i.e., the global graph summaries vary from node to node based on its associated cluster structures. For the nodes that share membership under a common set of clusters, the structure-aware graph summaries are similar. That makes the universal discriminator more powerful for discriminating the local and global patch pair representations from the false pairs across the relations. 
\par Here, we keep SSDCM's cluster learning component intact and use various pooling strategies to train the discriminator. The discriminator, thus, does not have any relation to the learned clusters and uses a common global summary paired with each node. In Table~\ref{table:cluster_pooling_novelty}, we see, structure-aware pooling methods are beneficial.  
\emph{We see that the pooling variants with SSDCM have better performance than DMGI's best-reported performance (Table 3), mainly due to learning of the clusters and using advanced pooling strategies} in place of DMGI's mean-pooling. Empirically, we observe that SSDCM with the alternative pooling variants struggle to converge consistently. However, SSDCM with a cluster-based graph summary does not suffer from similar convergence issues. \emph{Our proposed architecture in the last row outperforms all the candidate pooling techniques significantly on every dataset, depicting the effectiveness of our cluster-aware graph summary representations.}

\subsection{Effect of various regularizations} \label{section:cross}
\begin{table}[h] \centering
\begin{adjustbox}{max width=0.750\textwidth}
\begin{tabular}{@{}l|cc|cc|cc@{}}
\toprule
\multirow{2}{*}{Comparison} & \multicolumn{2}{c}{IMDB-MC}       & \multicolumn{2}{c}{FLICKR}        & \multicolumn{2}{c}{ACM}           \\ \cmidrule(l){2-7} 
                            & Micro-F1        & NMI-N           & Micro-F1        & NMI-N           & Micro-F1        & NMI-N           \\ \midrule
SSDCM                       & \textbf{67.796} & \textbf{0.22325} & \textbf{96.261} & \textbf{0.82171} & \textbf{88.324} & \textbf{0.94650} \\ \hline
SSDCM$-$cross                     & 66.613          & 0.20451          & 94.182          & 0.79671          & 86.371          & 0.91611          \\ 
SSDCM$-$cons                  & 66.069          & 0.20138          & 91.836           & 0.73658          & 84.889          & 0.88139          \\ 
SSDCM$-$(cons+cross)                  & 64.971          & 0.18735          & 88.374           & 0.71629          & 83.961          & 0.85420          
\\ \bottomrule
\end{tabular}
\end{adjustbox}  \caption{Effect of cross and consensus regularizations\\
\scriptsize{'+' and '-' sign denote augmentation or elimination of the components followed.}
}
\label{table:cross}
\end{table}

Here we compare the results of SSDCM without cross regularization with SSDCM to understand the influence of this factor. \emph{We see that removing cross-edge based regularization from the layer-wise node embeddings degrades the performance of the SSDCM considerably,} especially on FLICKR and ACM. Next, we verify the usefulness of learning a final consensus node embedding from the attention-aggregated positive and corrupted node embeddings. Recall that our universal discriminator learns to discriminate between true local--global patches from the corrupted ones with the intuition that the corrupted embeddings seek to improve the discriminative power of the resulting embeddings.  
\emph{We see that the consensus regularization indeed plays an essential role in enriching the final node embeddings -- an observation similar to DMGI's.} From Table~\ref{table:cross}, we see that the consensus embeddings improve the performance of Micro-F1 scores by a maximum of $4.425\%$ on FLICKR, followed by $3.435\%$, $1.558\%$ improvements on ACM, IMDB-MC respectively. 
\emph{SSDCM$-$(cons+cross) gives the worst performance among all the compared variations.} The reasons behind this are self-explanatory -- a) no cross-edges to align the relational node representations to each-other, b) it lacks in a discriminative capacity.

\section{Conclusion}
In this study, we propose a semi-supervised framework for representation learning in multiplex networks. This framework incorporates a unique InfoMax based learning strategy to maximize the MI between local and contextualized global graph summaries for effective joint modeling of nodes and clusters. Further, we use the cross-layer links to impose further regularization of the embeddings across the various layers of the multiplex graph.  Our novel approach, dubbed SSDCM, improves over the state-of-the-art over a wide range of experimental settings and four distinct downstream tasks, namely, classification, clustering, visualization, and similarity search, demonstrating the proposed framework's overall effectiveness.  In the future, we hope to extend this work in a couple of ways. First, we hope to improve the scalability of the approach -- perhaps by leveraging a graph coarsening and refinement strategy\cite{liang2021MILE} within SSDCM. Second, we propose to see if the ideas we have presented can be generalized for other types of multi-layer graphs (i.e., not just multiplex networks).

\bibliographystyle{unsrt}
\bibliography{main}

\clearpage
\appendix
\section{Appendix} \label{section:appendix}
\subsection{Ablation study} \label{section:ablation}
\begin{figure}[h] \centering
\minipage{0.475\textwidth}
  \includegraphics[width=\linewidth, keepaspectratio]{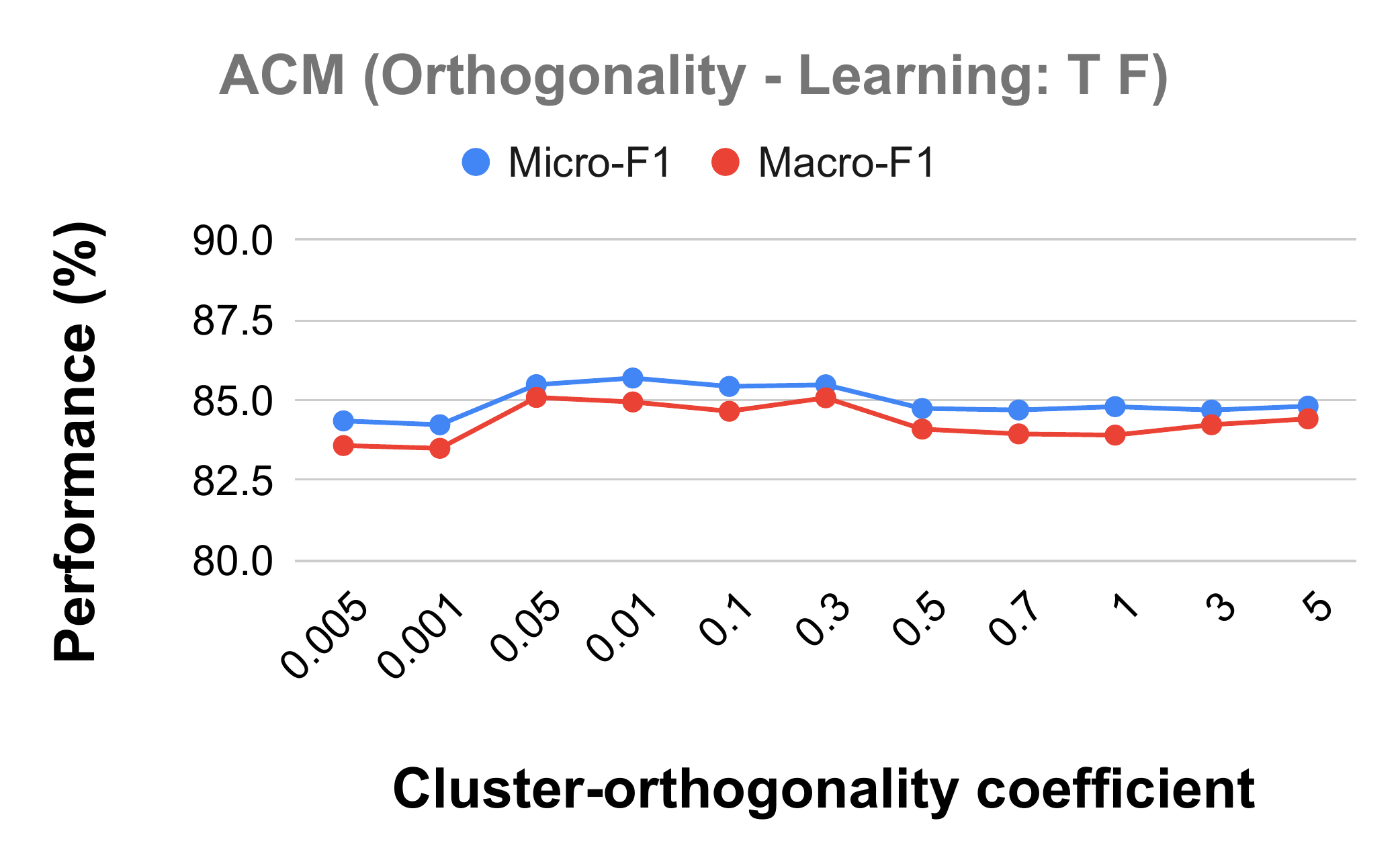}
\endminipage 
\minipage{0.475\textwidth}
  \includegraphics[width=\linewidth, keepaspectratio]{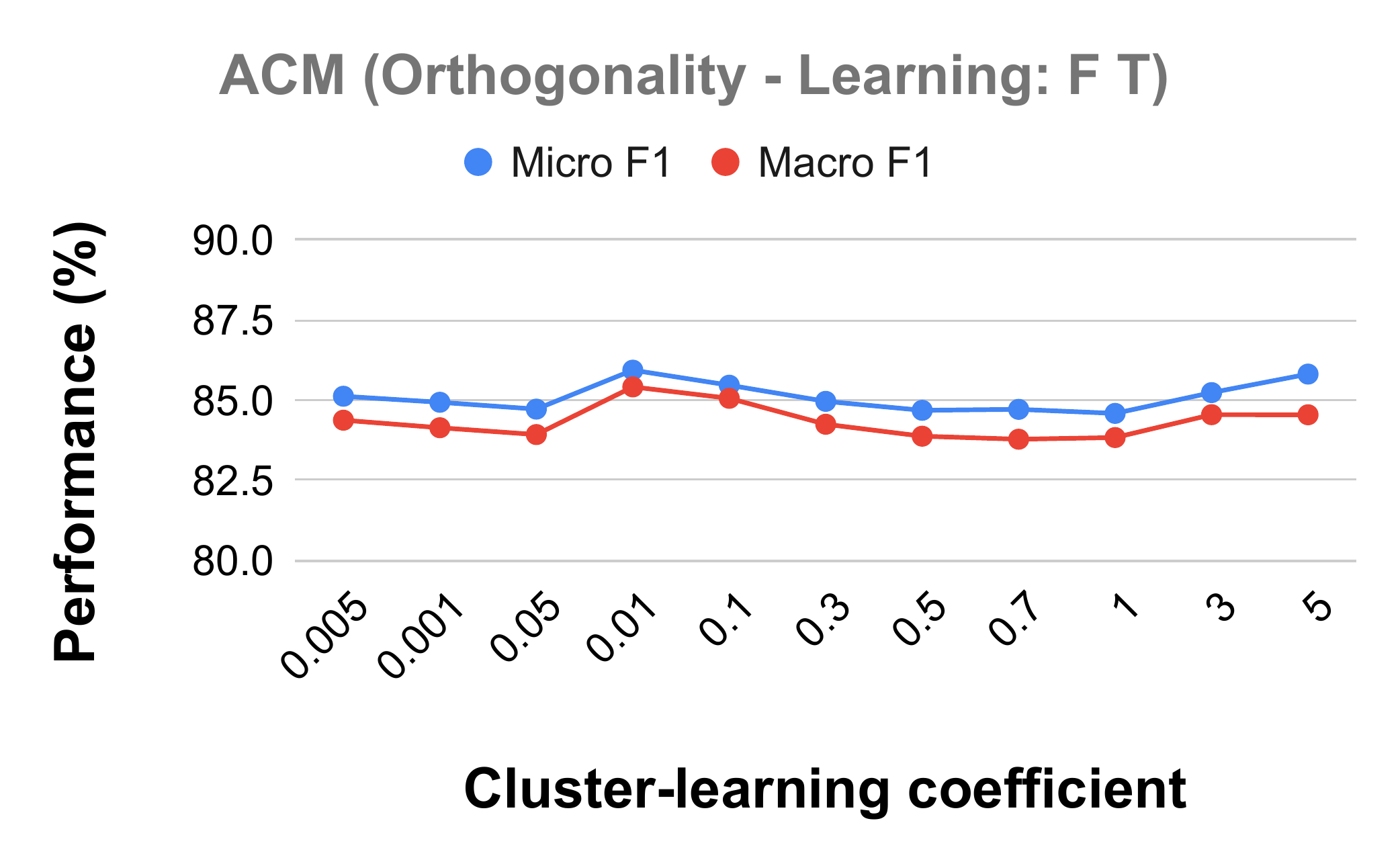}
\endminipage 
\caption{Ablation study of cluster learning components}\label{figure:ablation}
\end{figure}
\begin{table}[h] \centering
\begin{adjustbox}{max width=0.6750\textwidth}
\begin{tabular}{@{}c|lcccc@{}}
\toprule
\multicolumn{2}{c}{Variants}        & Micro F1        & Macro F1        & NMI-N          & NMI-C          \\ \midrule
\multicolumn{2}{c}{SSDCM (OL : FT)} & {\ul 85.584}    & {\ul 84.83}     & {\ul 0.889}    & {\ul 0.518}    \\
\multicolumn{2}{c}{SSDCM (OL : TF)} & 84.795          & 83.907          & 0.868          & 0.391          \\
\multicolumn{2}{c}{SSDCM (OL : TT)} & \textbf{88.324} & \textbf{88.571} & \textbf{0.947} & \textbf{0.651} \\ \bottomrule
\end{tabular} \end{adjustbox}  \caption{Impact of various cluster learning components}
\subcaption*{\scriptsize{Symbol meanings -- A: cluster assignment, L: cluster learning, O: cluster orthogonality. T: True, F: False -- denotes absence or presence of respective terms.}} \label{table:ablation}
\end{table}
In Table~\ref{table:ablation}, we study the impact of cluster related terms on the end-task performances by removing the relevant terms in two binary combinations. In OL: FT and OL: TF configurations, we remove the cluster orthogonality term and the semi-supervised cluster learning term, respectively. Removing the cluster learning term significantly impacts the NMI N and C scores by reducing the performance by $0.079$ and $0.26$ points. This configuration moderately affects the F1 scores. Removing the orthogonality term affects the classification performances with $2.74\%, 3.771\%$ reductions in Micro and Macro F1 scores. These reductions are less than the performance drops gotten from removing the cluster learning term in the case of F1 scores but still play a significant role.  \emph{The cluster learning term is seen to be more useful than the cluster orthogonality term for learning the cluster membership matrix.}

\par In Figure~\ref{figure:ablation}, we consider two possible combinations, namely, cluster assignment (Eqn~\ref{eqn:clust-assign})--learning (Eqn~\ref{eqn:clust-LapSmooth})--orthogonality (Eqn~\ref{eqn:clust-orth}) as LO: FT and TF (T: True, F: False), for dissecting the cluster learning objective. We perform a range search to see under which settings the best classification performances is achieved by varying a particular cluster related term in a range while removing or keeping the rest of the terms intact. The terms are varied in range of $\in \{0.005, 0.001, 0.05, 0.01, 0.1, 0.3, 0.5, 0.7, 1, 3, 5\}$. In FT configuration, cluster orthogonality is varied in the absence of the cluster learning term. It gives best performance for values $\in \{0.05, 0.3\}$. Over a higher range of values, the performances become less fluctuating. In TF configuration, cluster learning is varied in the absence of cluster orthogonality. At $0.01$ it gives the best performance in terms of Micro and Macro F1 for ACM. Again, an upward trend in performances can be seen for values $\in [1-5]$.

\subsection{Varying number of clusters} \label{section:clusters}
\begin{figure}[h!] \centering
\minipage{0.475\textwidth}
  \includegraphics[width=\linewidth, keepaspectratio]{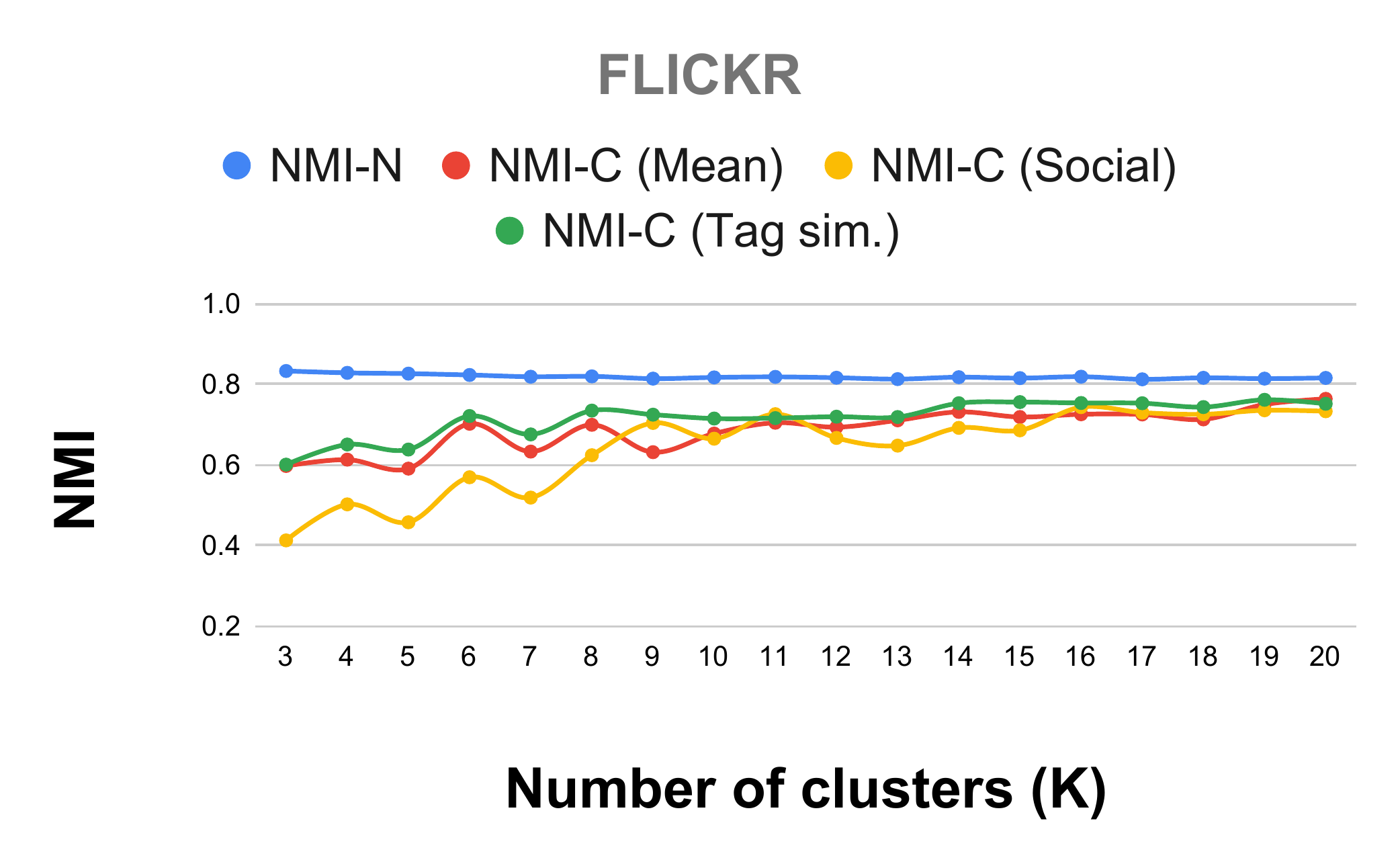}
\endminipage 
\minipage{0.475\textwidth}
  \includegraphics[width=\linewidth, keepaspectratio]{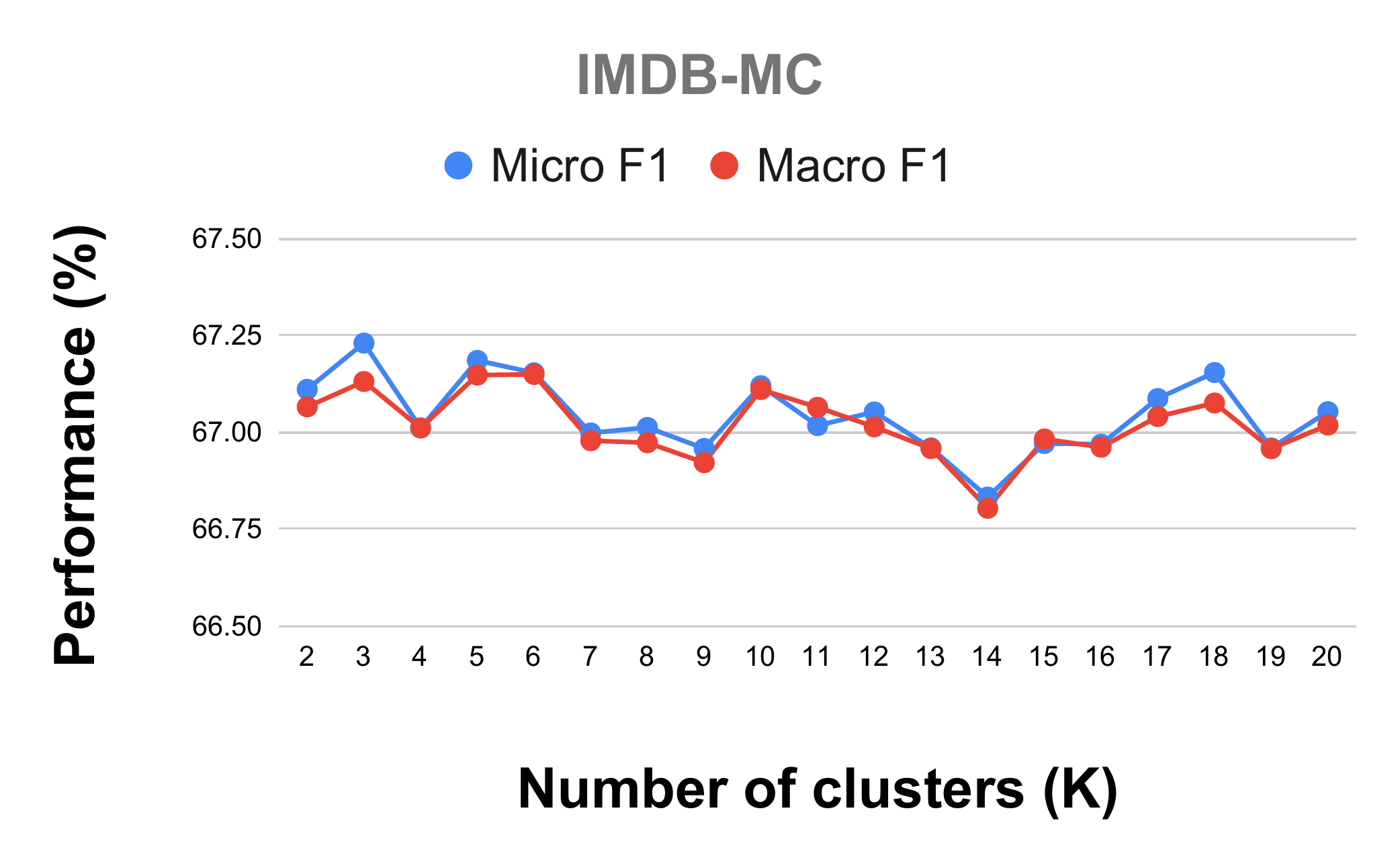}
\endminipage \\
\minipage{0.475\textwidth}
  \includegraphics[width=\linewidth, keepaspectratio]{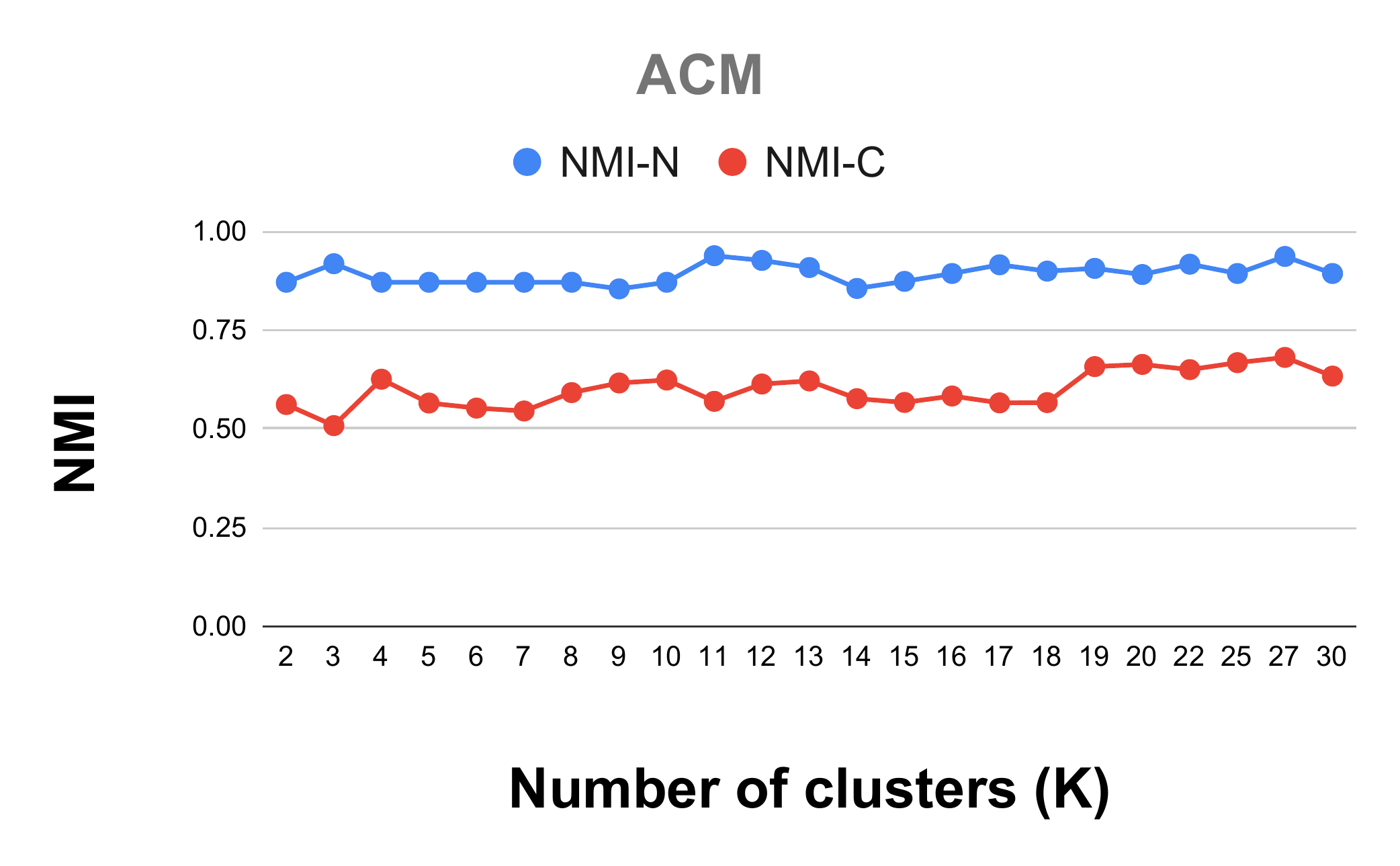}
\endminipage 
\minipage{0.475\textwidth}
  \includegraphics[width=\linewidth, keepaspectratio]{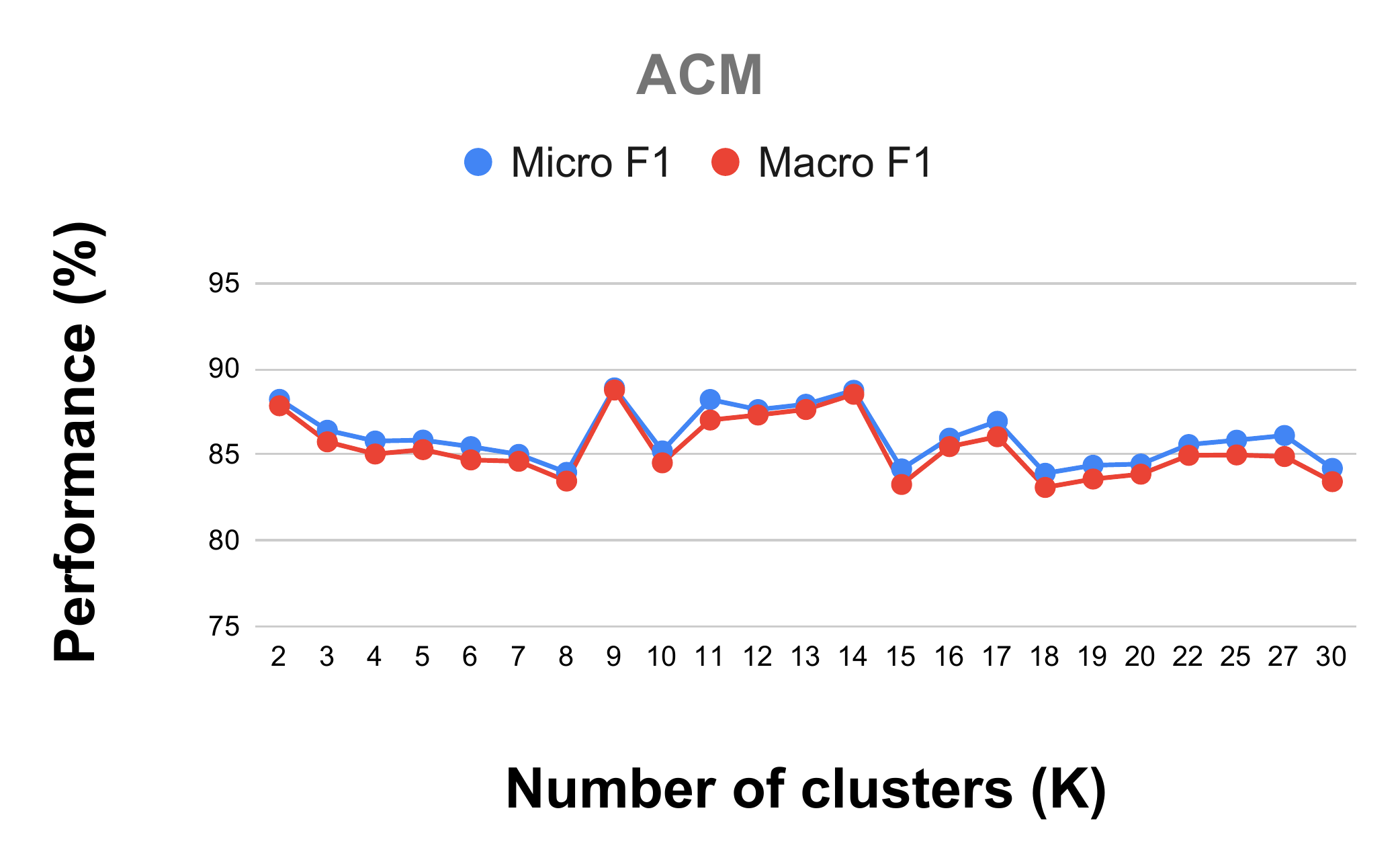}
\endminipage 
\caption{Varying number of clusters}\label{figure:clusters}
\subcaption*{\scriptsize{Number of clusters K is varied for FLICKR, IMDB-MC and ACM. i) Micro, Macro F1 scores (on right), ii) NMI using node embeddings and cluster memberships (on left) are plotted. Best performances of DMGI (no cluster learning) are -- a) for FLICKR, NMI-N: $0.644$, b) for IMDB-MC, Micro-F1: $65.210$, Macro-F1: $65.066$, and c) for ACM, Micro-F1: $81.205$, Macro-F1: $80.802$, NMI-N: $0.837$}}
\end{figure}
Here we study SSDCM's sensitivity towards varying the number of clusters $K$. We also verify whether there is a need to learn the cluster structures at all. We take the optimal hyper-parameter combination and vary the number of clusters in the range $[2-20]$ and $[2-30]$ for FLICKR, IMDB-MC, ACM, respectively. \emph{Compared to DMGI's best performance scores, clear differences can be seen in Figure~\ref{figure:clusters} for SSDCM that speaks to the effectiveness of learning clusters to enrich node embeddings.} 
\par We plot the NMI-N and NMI-C scores (mean and layer-wise cluster memberships) while varying $ K $ for FLICKR. We see less perturbation in NMI-N scores than NMI-C scores here. As $K$ goes higher, the layer-wise and mean cluster membership based NMI scores increase before flattening at $K=20$.
For IMDB-MC, We can see Micro F1 scores are best at $K=Q=3$, i.e., when the number of classes and clusters are the same. For ACM, varying $K$ improves Micro and Macro F1 scores at $K \in \{2,3\} < (Q=5)$, i.e., when SSDCM learns high-level clusters. Even when $K \geq Q$, i.e., when SSDCM learns small clusters of same class data. We see a gradual improvement in both the NMI scores for ACM when $k\in [9-30]$. NMI-N and NMI-C tend to give different NMI scores. The possible interpretation of this performance difference lies in the fact that -- in NMI-N, the K-Means algorithm applied to the node embeddings of considerable hidden dimensions $(d=64)$ and the NMI scores are calculated for ground truth clusters. Whereas, cluster memberships $H$ are of comparatively low dimensions $(K)$, and in NMI-C, we directly use the learned cluster membership probabilities to derive the NMI scores.

\subsection{Reproducibility}\label{subsubsection:reproducibility}
\subsubsection{Baselines.}\label{subsubsection:baselines_detailed}
In Table~\ref{table:hyperparam_range}, we give the details of hyper-parameter range search for all the competing methods --- which is self-explanatory.
\begin{table*}[h] \centering
\begin{adjustbox}{max width=0.9\textwidth}
\begin{tabular}{@{}l|llllllllll@{}}
\toprule
                 & \textbf{Nodes} & \textbf{Layers} & \textbf{Node Types} & \textbf{\begin{tabular}[c]{@{}l@{}}Intra-Layer \\ Relations\end{tabular}} & \textbf{Edges} & \textbf{Features} & \textbf{Weighted} & \textbf{Directed} & \textbf{Multi-Class} & \textbf{Classes}   \\ \midrule
\textbf{ACM}~\cite{wang2019heterogeneous}     & 7427           & 5               & \textbf{\textsc{Paper (P)}}           & PAP                                                                       & 118453         & 767               & True              & False             & True                  & 5                  \\
\textbf{}        &                &                 & Author (A)          & PAIAP                                                                     & 8353678        & Paper Title       &                   &                   &                       & Conference (C)     \\
\textbf{}        &                &                 & Proceeding (V)      & PSP                                                                       & 14997105       & \& Abstract       &                   &                   &                       &                    \\
\textbf{}        &                &                 & Institute (I)       & PVP                                                                       & 1048129        &                   &                   &                   &                       &                    \\
\textbf{}        &                &                 & Subject (S)         & PP                                                                        & 19324          &                   &                   &                   &                       &                    \\
\hline
\textbf{DBLP}~\cite{wang2019heterogeneous}    & 4057           & 4               & \textbf{\textsc{Author (A)}}          & APA                                                                       & 11113          & 8920              & True              & False             & True                  & 4                  \\
\textbf{}        &                &                 & Paper (P)           & APAPA                                                                     & 40703          & Paper Title       &                   &                   &                       & Research Field (F) \\
\textbf{}        &                &                 & Conference (C)      & APCPA                                                                     & 5000495        & \& Abstract       &                   &                   &                       &                    \\
\textbf{}        &                &                 & Term (T)            & APTPA                                                                     & 12924399       &                   &                   &                   &                       &                    \\
\hline
\textbf{SLAP}~\cite{zhang2018deep}    & 20419          & 6               & \textbf{\textsc{Gene (G)}}            & GPG                                                                       & 832924         & 2695              & True              & False             & True                  & 15                 \\
\textbf{}        &                &                 & Gene Ontology (O)   & GTG                                                                       & 606974         & Gene Ontology     &                   &                   &                       & Gene Family (F)    \\
\textbf{}        &                &                 & Pathway (P)         & GDCDG                                                                     & 36190          & Description       &                   &                   &                       &                    \\
\textbf{}        &                &                 & Compound (C)        & GOG                                                                       & 6371558        &                   &                   &                   &                       &                    \\
\textbf{}        &                &                 & Tissue (T)          & GDG                                                                       & 14988          &                   &                   &                   &                       &                    \\
\textbf{}        &                &                 & Disease (D)         & GG                                                                        & 344496         &                   &                   &                   &                       &                    \\
\hline
\textbf{IMDB-MC}~\cite{park2020unsupervised} & 3550           & 2               & \textbf{\textsc{Movie (M)}}           & MAM                                                                       & 66428          & 2000              & True              & False             & True                  & 3                  \\
\textbf{}        &                &                 & Actor (A)           & MDM                                                                       & 13788          & Movie Plot        &                   &                   &                       & Movie Genre (G)          \\
\textbf{}        &                &                 & Director (D)        &                                                                           &                & \& Summary        &                   &                   &                       &                    \\
\hline
\textbf{IMDB-ML}~\cite{pham2017column} & 18352          & 3               & \textbf{\textsc{Movie (M)}}           & MAM                                                                       & 1455381        & 1000              & True              & True              & False                 & 9                  \\
\textbf{}        &                &                 & Actor (A)           & MDM                                                                       & 923173         & Movie Plot        &                   &                   &                       & Movie Genre (G)          \\
\textbf{}        &                &                 & Director (D)        & MEM                                                                       & 127243         & \& Summary        &                   &                   &                       &                    \\
\textbf{}        &                &                 & Actress (E)         &                                                                           &                &                   &                   &                   &                       &                    \\
\hline
\textbf{FLICKR}~\cite{luo2020deep}  & 10364          & 2               & \textbf{\textsc{User (U)}}            & Friendship                                                                & 390938         & NA                & True              & True              & True                  & 7                  \\
\textbf{}        &                &                 &                    & Tag-similarity                                                            & 115113         &                   &                   &                   &                       & Social Group (G)   \\  \hline
\textbf{AMAZON}~\cite{park2020unsupervised}  & 17857  & 3      & \textbf{\textsc{Product (P)}}       & Co-purchase           & 1501401 & 2395                      & True     & False    & True         & 5                  \\
        &      &        &                & Co-view               & 590961                       & Product       &          &          &              & Product Category (C)       \\
        &    &        &                  & Similar               & 102027  & Description                         &          &          &              &                    \\
\bottomrule
\end{tabular}
\end{adjustbox}
\caption{Dataset statistics} \label{table:dataset}
\end{table*}

\begin{table*}[h] \centering
\begin{adjustbox}{max width=0.95\textwidth}
\begin{tabular}{@{}l|llllllllll@{}}
\toprule
\textbf{Methods}                 & \multicolumn{10}{c}{\textbf{Experiment setup \& hyper-parameter range}}                                                                                                                                                \\ \midrule
\textbf{\textsc{HAN}}~\cite{wang2019heterogeneous}                    & \multicolumn{10}{l}{l2 coefficient=\{0.0001, 0.0005, 0.001, 0.005\}, learning rate=\{0.0001, 0.0005, 0.001, 0.005\}, attention heads=\{1,2,4,8\}, metapath attention dimension=128}      \\
\textbf{\textsc{MGCN}}~\cite{ghorbani2019mgcn}                   & \multicolumn{10}{l}{network \& label coefficient=\{0.01, 0.1, 1.0, 10.0\}, l2 coefficient=\{0.0005, 0.005\}, learning rate=\{0.0005, 0.001, 0.05, 0.01\}, stacked GCNs=2}                \\
\textbf{\textsc{RGCN}}~\cite{schlichtkrull2018modeling}                   & \multicolumn{10}{l}{l2 coefficient=\{0.0005, 0.005\}, learning rate=\{0.0005, 0.001, 0.05, 0.01\}, no of bases=no of relations, number of hidden layers=2}                               \\
\textbf{\textsc{GUNets}}~\cite{gao2019graph}           & \multicolumn{10}{l}{l2 coefficient=\{0.0001, 0.001\}, learning rate=\{0.01, 0.05, 0.001, 0.0005\}, depth=\{3, 4, 5\}, pool ratio=\{0.2, 0.4, 0.6, 0.8\}}                                 \\
\textbf{\textsc{DMGC}}~\cite{luo2020deep}                   & \multicolumn{10}{l}{network coefficient=\{1.0, 0.8, 0.6, 0.4\}, cross reg.=\{0.2, 0.4, 0.6, 0.8\}, l2 coef=0.0001, learning rate=\{0.0005, 0.001, 0.05, 0.01\}, stacks in AutoEncoder=2} \\
\textbf{\textsc{DMGI}}~\cite{park2020unsupervised}                   & \multicolumn{10}{l}{network \& label coefficient=\{0.001,  0.01, 0.1, 1.0\}, l2 coefficient=\{0.0001,  0.001\}, learning rate=\{0.0001, 0.0005, 0.001, 0.005\}}                          \\
\multirow{2}{*}{\textbf{\textsc{SSDCM}}} & \multicolumn{10}{l}{\multirow{2}{*}{network, label \& cluster coefficient=\{0.001,  0.01, 0.1\}, l2 coefficient=0.0001, cross regularization=0.001, learning rate=\{0.0001, 0.0005, 0.001, 0.005\}}}   \\
                       & \multicolumn{10}{l}{}      \\                                                                      \bottomrule  
\specialrule{2.5pt}{2pt}{2pt}
\multirow{2}{*}{Default to All} & \multicolumn{8}{l}{hidden units=64, epochs=10000, patience=20, attention heads=2, non-linearity=prelu, no of clusters=no of classes, $\epsilon=3.0$, GCN layers $=2$,}  \\
                                & \multicolumn{8}{l}{validation set based hyperparameter tuning, features=adjacency for non-attributed graphs, no node aggregation strategy=mean-pooling.}
     \\
\hline
\end{tabular} \end{adjustbox} \caption{Experiment setup \& hyper-parameter range search for competing methods}
\label{table:hyperparam_range}
\end{table*}

\subsubsection{Datasets.}\label{subsubsection:datasets_detailed}
Here in Table~\ref{table:dataset}, we provide the detailed statistics of the datasets used for evaluation. 
We have used two versions of the IMDB dataset, one multi-class version \textbf{\textsc{IMDB-MC}} as used in DMGI, and, another multi-label version \textbf{\textsc{IMDB-ML}} from the Column Networks (CLN)~\cite{pham2017column}. In both versions, movie features are extracted from movie plot summary with movie genres as functional classes. We used multiplex versions of bibliographic datasets ACM and DBLP. For \textbf{\textsc{ACM}}~\cite{wang2019heterogeneous}, we extracted papers of five conferences \footnote{Conferences = ['KDD', 'WWW', 'SIGIR', 'SIGMOD', 'CIKM']} and created a multiplex network that includes layers of paper nodes connected by co-authors, similar subjects, similar venues, co-authors belonging from the same institutes, and citation relationships. Here, the task is to classify them according to the conferences as they are published. \textbf{\textsc{DBLP}}~\cite{wang2019heterogeneous} is a multiplex network of authors. The authors are classified by their field of research-interests \footnote{Fields = [Data Mining (DM), Artificial Intelligence (AI), Computer Vision (CV), Natural language Processing (NLP)].}. In both the bibliographic datasets, the terms extracted from the paper title and abstract are used as local features for the nodes under consideration. In \textbf{\textsc{SLAP}} \cite{kong2012meta, zhang2018deep}, multiple layers of interactions characterize a gene --- including tissue-specific, biological pathways involved, disease associations,  phylogenetic profile, gene expression, chemicals involved to treat associated diseases, etc. Each gene has ontology related terms associated with it as attributes, and it can belong to any of the most frequently occurring fifteen gene Families (F). We have \textbf{\textsc{AMAZON}}~\cite{he2016ups, park2020unsupervised}, which is originally multiplex in nature, i.e., the multiplexity is not inferred from composite relations. This network is extracted from the product review metadata of the Amazon website. Target instances, i.e., products exhibit also-bought, also-viewed, and similar-to -- three layers of relations among them. Most frequently occurred terms are extracted from product title as node features. The task is to classify the products into any product categories \footnote{Product Categories in AMAZON Multiplex Network = ['Appliances', 'Automotive', 'Patio Lawn \& Garden', 'Pet Supplies', 'Home \& Kitchen']}. \textbf{\textsc{FLICKR}}~\cite{luo2020deep} is a non-attributed multiplex social network of users (U) who belong to various communities of interest. It has a friendship layer and a tag similarity-based connection layer among the users. A user is categorized based on their membership to any of the social groups. In all the datasets mentioned in Table~\ref{table:dataset}, cross-layer edges link two nodes in different layers if they refer to the same node.

\end{document}